\documentclass[10pt,twocolumn,letterpaper]{article}

\usepackage{iccv}
\usepackage{times}
\usepackage{epsfig}
\usepackage{graphicx}
\usepackage{amsmath}
\usepackage{amssymb}

\usepackage[accsupp]{axessibility}
\usepackage{algorithm}
\usepackage{algorithmic}
\usepackage{tabularx}
\usepackage{multirow}
\usepackage{hhline}
\newcolumntype{Y}{>{\centering\arraybackslash}X}

\makeatletter
\def\thickhline{%
  \noalign{\ifnum0=`}\fi\hrule \@height \thickarrayrulewidth \futurelet
   \reserved@a\@xthickhline}
\def\@xthickhline{\ifx\reserved@a\thickhline
               \vskip\doublerulesep
               \vskip-\thickarrayrulewidth
             \fi
      \ifnum0=`{\fi}}
\makeatother
\newlength{\thickarrayrulewidth}
\newcolumntype{?}{!{\vrule width 1pt}}

\usepackage[breaklinks=true,bookmarks=false]{hyperref}

\iccvfinalcopy 

\def\iccvPaperID{6638} 

\ificcvfinal\fi

\begin{document}

\newcommand{\modelname}{MAPConNet}
\title{\modelname: Self-supervised 3D Pose Transfer with Mesh and Point Contrastive Learning}


\author{Jiaze Sun$^1$\thanks{Corresponding author email: \url{j.sun19@imperial.ac.uk}.} 
\quad Zhixiang Chen$^{2}$ \quad Tae-Kyun Kim$^{1,3}$ \\
{\small $^1$Imperial College London \quad $^2$University of Sheffield \quad $^3$Korea Advanced Institute of Science and Technology}}

\maketitle
\ificcvfinal\thispagestyle{empty}\fi

\begin{abstract}
    3D pose transfer is a challenging generation task that aims to transfer the pose of a source geometry onto a target geometry with the target identity preserved. Many prior methods require keypoint annotations to find correspondence between the source and target. Current pose transfer methods allow end-to-end correspondence learning but require the desired final output as ground truth for supervision. Unsupervised methods have been proposed for graph convolutional models but they require ground truth correspondence between the source and target inputs. We present a novel self-supervised framework for 3D pose transfer which can be trained in unsupervised, semi-supervised, or fully supervised settings without any correspondence labels. We introduce two contrastive learning constraints in the latent space: a mesh-level loss for disentangling global patterns including pose and identity, and a point-level loss for discriminating local semantics. We demonstrate quantitatively and qualitatively that our method achieves state-of-the-art results in supervised 3D pose transfer, with comparable results in unsupervised and semi-supervised settings. Our method is also generalisable to unseen human and animal data with complex topologies\footnote[2]{Code: \url{https://github.com/justin941208/MAPConNet}.}.
\end{abstract}
\section{Introduction}
\label{sec:introduction}
\begin{figure}[t]
\begin{center}
\includegraphics[width=0.99\linewidth]{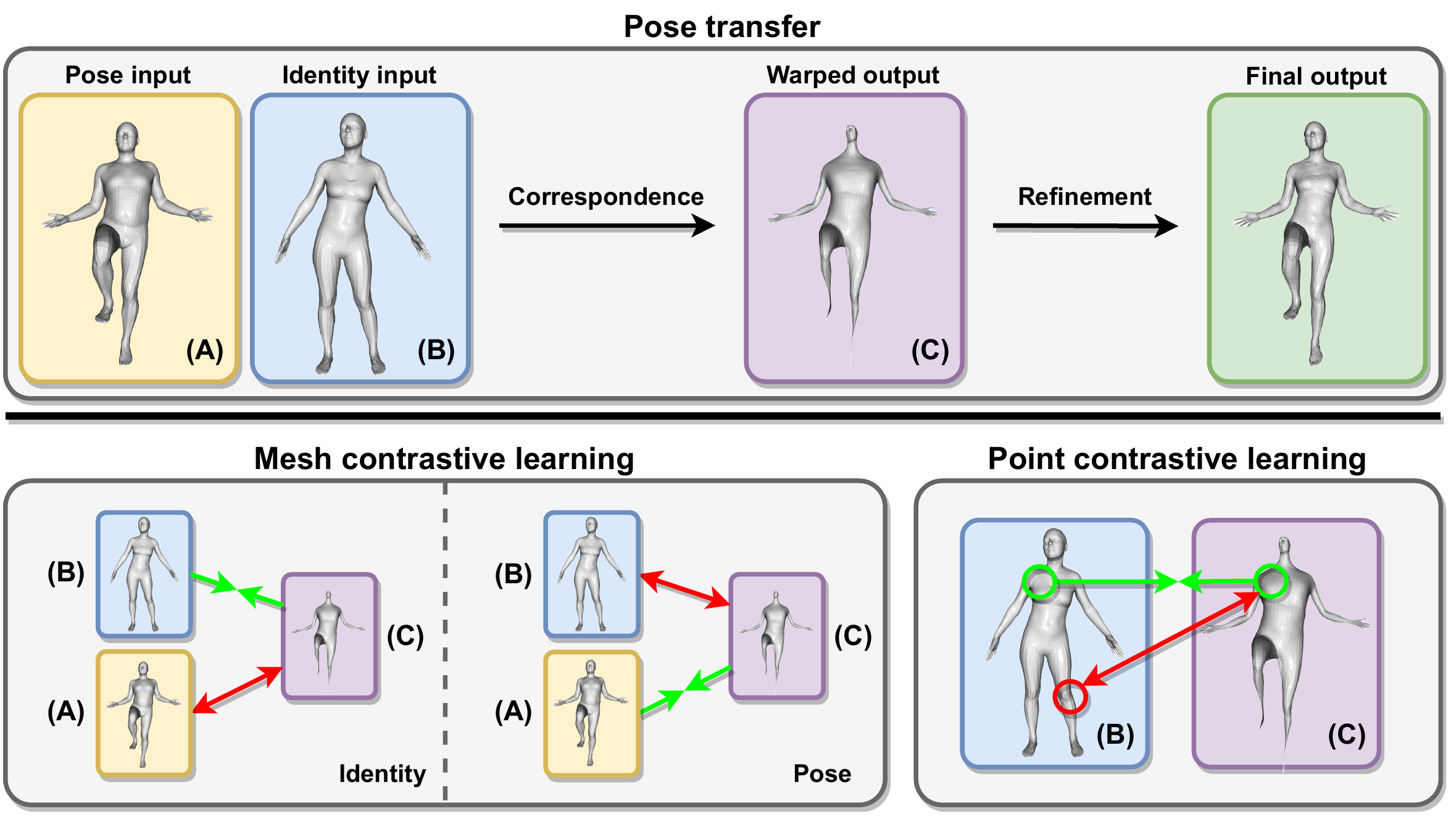}
\end{center}
  \caption{\textbf{Overview of our framework.} Top: our pose transfer pipeline. Bottom: our contrastive learning scheme, with a mesh-level loss for disentangling global pose and identity and a point-level loss for discriminating local semantics.}
\label{fig:teaser}
\end{figure}

3D pose transfer \cite{Sumner2004Deformation,Wang2020Neural,zhou2020unsupervised,song20213d} is a challenging generation task in which the pose of a source geometry is transferred to a target geometry whilst preserving the identity of the target geometry (see top half of Figure \ref{fig:teaser}). This has many potential applications in areas including animation, human modelling, virtual reality, and more. It also provides a more affordable way of generating synthetic 3D data which can be expensive to produce in the real world.

One of the main challenges of 3D pose transfer is that current methods still put certain requirements on their training data making it difficult to collect and expensive to annotate. One requirement is having correspondence labels, which are pairs of vertices that correspond to each other semantically between two point clouds or meshes. Many prior pose transfer methods either require ground truth correspondence \cite{Sumner2004Deformation,Ben2009Spatial,Baran2009Semantic,Yang2018Biharmonic,zhou2020unsupervised} or simply neglect the issue \cite{Wang2020Neural}. Requiring ground truth correspondence is costly, and neglecting correspondence would adversely impact model performance. \cite{song20213d} proposed learning a correspondence module based on optimal transport in an end-to-end fashion through the pose transfer task. However, their method is supervised and requires the mesh with the desired pose and identity as ground truth. This puts another requirement on the dataset: having multiple subjects performing exactly the same set of poses, which is unfeasible. \cite{zhou2020unsupervised} proposed an unsupervised approach for registered meshes which only requires each subject to perform multiple poses and they do not have to align exactly across subjects -- a more practical requirement for real datasets \cite{Bogo2017DFAUST,Mahmood2019AMASS}. However, their approach is based on graph convolutional networks (GCNs) and ARAP deformation which require ground truth correspondence.

We propose a self-supervised framework (Figure \ref{fig:teaser}) for 3D pose transfer with \emph{Mesh And Point Contrastive learning}, \emph{\modelname}, requiring no correspondence labels or target outputs with the desired pose and identity as ground truth for supervision. It can be applied in supervised, unsupervised, and semi-supervised settings, and does not need the pose and identity inputs to have the same ordering or number of points. We propose to adapt the unsupervised approach by \cite{zhou2020unsupervised} for pose transfer on unaligned meshes, using \cite{song20213d} as our baseline. To prevent the network from exploiting shortcuts in unsupervised learning that circumvent the desired objective, we introduce disentangled latent pose and identity representations. To strengthen the disentanglement and guide the learning process more effectively, we propose \emph{mesh-level contrastive learning} to force the model's intermediate output to have matching latent identity and pose representations with the respective inputs. To further improve the quality of the model's intermediate output as well as the correspondence module, we also propose \emph{point-level contrastive learning}, which enforces similarity between representations of corresponding points and dissimilarity between non-corresponding ones. In summary:
\begin{itemize}
    \item We present \emph{\modelname}, a novel self-supervised network for 3D pose transfer with \emph{Mesh And Point Contrastive} learning. Our model requires no ground truth correspondence or target outputs for supervision.
    \item We introduce two levels of contrastive learning constraints, with a mesh-level loss for global disentanglement of pose and identity and a point-level loss for discrimination between local semantics.
    \item We achieve state-of-the-art results through extensive experiments on human and animal data, and demonstrate competitive and generalisable results in supervised, unsupervised, and semi-supervised settings.
\end{itemize}
\section{Related work}
\label{sec:related_work}

\textbf{Deep learning on 3D data.} Recent years have seen a surge in deep learning methods on 3D data such as point clouds, meshes and voxels. \cite{Wu2015ShapeNets} and \cite{Maturana2015VoxNet} operate on voxels using 3D convolutions which would be computationally expensive for high-dimensional data such as ours. \cite{feng2019meshnet} and \cite{Tan2018Variational} are designed for meshes but include fully-connected layers which are memory-intensive. Some GCN models \cite{Ranjan2018COMA,Ma2020CAPE,zhou2020unsupervised} include down- and up-sampling layers and others \cite{Lim2019Spiral,gong2019spiralnet++} incorporate novel operations, both of which improve efficiency. However, all their architectures rely on a template structure to implement and is not suitable in our scenario. As for point clouds, \cite{Qi2017pointnet} and \cite{Qi2017PointNet++} use shared weights across points and adopt aggregation strategies to enforce order-invariance. We use shared weights without aggregation to preserve detailed identity information.

\textbf{3D pose/deformation transfer.} Pose transfer aims to transfer the pose of a source geometry to a target without changing the target's identity. Deformation transfer methods \cite{Sumner2004Deformation,Ben2009Spatial,Baran2009Semantic,Yang2018Biharmonic,Yifan2020Neural,liao2022pose} require a template pose across different identities and sometimes also correspondence labels, which are unavailable in our and most realistic settings. Image-to-image translation methods \cite{CycleGAN2017,choi2018stargan,He2019AttGAN,Park2019Semantic,Sun2020MatchGAN,Sun2022SeCGAN} have also been repurposed for pose transfer due to their relevance. \cite{Gao2018Automatic} used CycleGAN \cite{CycleGAN2017} for pose transfer but requires retraining for each new pair of identities. \cite{Basset2021Neural} reformulated the problem as ``identity transfer" but required vertex correspondence. \cite{Wang2020Neural} used SPADE normalisation \cite{Park2019Semantic} to inject target identity into the source. \cite{song20213d} added correspondence learning to \cite{Wang2020Neural} and improved the normalisation method. However, both require ground truth outputs. \cite{zhou2020unsupervised} proposed an unsupervised framework but requires correspondence labels. We do not require ground truth outputs or correspondence labels. \cite{song2023unsupervised} proposed a dual reconstruction objective in a similar spirit to \cite{zhou2020unsupervised} to enable unsupervised learning in \cite{song20213d}. In contrast, our approach is to force the model to learn disentangled latent pose and identity codes and impose mesh- and point-level contrastive losses on them, which improves performance in both supervised and unsupervised settings.

\textbf{Self-supervised learning on 3D data.} Self-supervised learning is the paradigm of automatically generating supervisory signals in training and has been successful in 2D \cite{Schroff2015FaceNet,doersch2015unsupervised,pathak2016context,zhang2016colorful,gidaris2018unsupervised,chen2019self,Sun2020MatchGAN}. Some 2D approaches have also been adopted for 3D data, such as rotation \cite{poursaeed2020self} and completion \cite{Wang2021Unsupervised}, but we choose contrastive learning as it can be easily adapted to suit our needs. Most existing contrastive learning approaches for 3D data \cite{Zhang2019Unsupervised,PointContrast2020,Sanghi2020Info3D,du2021self,Afham2022CrossPoint} focus on learning invariance across views or rigid transformations for scenes or simple objects, whereas we address more fine-grained patterns such as identity, pose, and correspondence for complex shapes including humans and animals.
\section{Methodology}
\label{sec:methodology}

We now present our problem setting and proposed {\modelname} in detail. Given a pose (i.e. source) mesh $\mathbf{x}^{A1}\in\mathbb{R}^{N_{pose}\times 3}$ and identity (i.e. target) mesh $\mathbf{x}^{B2}\in\mathbb{R}^{N_{id}\times 3}$, where the letters $A,B,\ldots$ and numbers $1,2,\ldots$ denote the identities and poses of the meshes respectively, our goal is to train a network $G$ to produce a new mesh $\hat{\mathbf{x}}^{B1} = G(\mathbf{x}^{A1}, \mathbf{x}^{B2})\in\mathbb{R}^{N_{id}\times 3}$ which inherits pose 1 and identity $B$. The integers $N_{pose}$ and $N_{id}$ are the numbers of vertices in the pose and identity meshes, respectively. The meshes are treated as point clouds by our model, but connectivity is required for training. In addition, we do not require the vertices of both inputs to share the same order, but the output mesh $\hat{\mathbf{x}}^{B1}$ would follow the same order as $\mathbf{x}^{B2}$.

\subsection{Preliminaries}

\begin{figure*}[t]
\begin{center}
\includegraphics[width=0.99\linewidth]{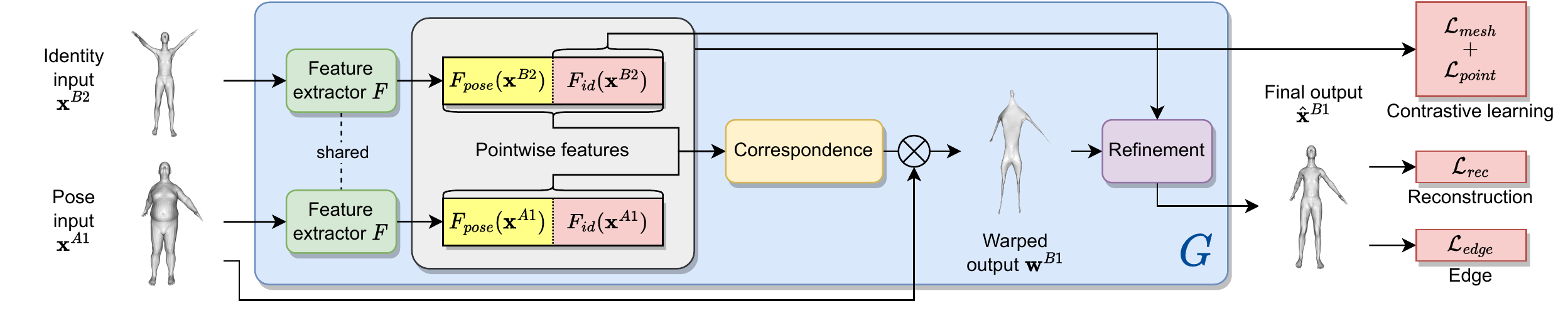}
\end{center}
  \caption{\textbf{Our supervised learning pipeline}. The feature extractor $F$ embeds the inputs $\mathbf{x}^{B2}$ and $\mathbf{x}^{A1}$ into a shared latent space. The correspondence module warps $\mathbf{x}^{A1}$ into $\mathbf{w}^{B1}$ which should have the same vertex order as $\mathbf{x}^{B2}$. In addition, we separate the latent codes into identity and pose, where only the identity channels are used as style conditioning to refine $\mathbf{w}^{B1}$ into the final output $\hat{\mathbf{x}}^{B1}$. We propose mesh and point contrastive learning on top of the existing reconstruction and edge losses. Our model is based on 3D-CoreNet \cite{song20213d}.}
\label{fig:pipeline}
\end{figure*}

For our baseline, we choose 3D-CoreNet \cite{song20213d} -- a prior state-of-the-art 3D pose transfer model that requires no ground truth correspondence.
It has two modules: correspondence and refinement. The correspondence module produces an intermediate ``warped'' output $\mathbf{w}^{B1}\in\mathbb{R}^{N_{id}\times 3}$ inheriting the pose from $\mathbf{x}^{A1}$ but the vertex order of $\mathbf{x}^{B2}$. Specifically, the warped output is obtained by $\mathbf{w}^{B1}=\mathbf{T}\mathbf{x}^{A1}$, where $\mathbf{T}\in\mathbb{R}_{+}^{N_{id}\times N_{pose}}$ is an optimal transport (OT) matrix learned based on the latent features of both inputs.
The refinement module then uses the features of the identity input $\mathbf{x}^{B2}$ as style condition for the warped output $\mathbf{w}^{B1}$, refining it through elastic instance normalisation and producing the final output $\hat{\mathbf{x}}^{B1}$.
The training of 3D-CoreNet is supervised: given the model output $\hat{\mathbf{x}}^{B1}$ and the ground truth output $\mathbf{x}^{B1}$, it minimises the reconstruction loss
\begin{equation}\label{eqn:rec_loss}
    \mathcal{L}_{rec}(\hat{\mathbf{x}}^{B1}; \mathbf{x}^{B1}) = \frac{1}{3N_{id}}\|\hat{\mathbf{x}}^{B1}-\mathbf{x}^{B1}\|_F^2,
\end{equation}
where $\|\cdot\|_F$ is the Frobenius norm. For this loss to work properly, the ground truth $\mathbf{x}^{B1}$ and identity input $\mathbf{x}^{B2}$ must have the same dimensions and vertex order. In addition to $\mathcal{L}_{rec}$, an edge loss is used to help generate smoother surfaces and prevent flying vertices \cite{Wang2020Neural,song20213d}. Given the model output $\hat{\mathbf{x}}^{B1}$ and identity input $\mathbf{x}^{B2}$, which should have matching vertex orders, the edge loss is given by
\begin{equation}
    \mathcal{L}_{edge}(\hat{\mathbf{x}}^{B1}; \mathbf{x}^{B2}) = \frac{1}{|\mathcal{E}|}\sum_{(j,k)\in\mathcal{E}} \left|\frac{\|\hat{\mathbf{x}}_j^{B1}-\hat{\mathbf{x}}_k^{B1}\|_2}{\|\mathbf{x}_j^{B2}-\mathbf{x}_k^{B2}\|_2} - 1 \right|,
\end{equation}
where $\mathcal{E}$ is the set of all index pairs representing vertices that are connected by an edge, and $\hat{\mathbf{x}}_j^{B1},\mathbf{x}_j^{B2}\in\mathbb{R}^3$ are the coordinates of the $j$-th (similarly, $k$-th) vertices of $\hat{\mathbf{x}}^{B1}$ and $\mathbf{x}^{B2}$ respectively.
Finally, the overall supervised loss is given by
\begin{equation}\label{eqn:sup}
    \mathcal{L}_{s} = \lambda_{rec}\mathcal{L}_{rec}(\hat{\mathbf{x}}^{B1}; \mathbf{x}^{B1}) + \lambda_{edge}\mathcal{L}_{edge}(\hat{\mathbf{x}}^{B1}; \mathbf{x}^{B2}),
\end{equation}
where $\lambda_{rec}$ and $\lambda_{edge}$ are the weights for the two losses.

\subsection{Latent disentanglement of pose and identity} \label{ssec:lat_dis}
Our supervised pipeline is shown in Figure \ref{fig:pipeline} with proposed loss terms $\mathcal{L}_{mesh}$ and $\mathcal{L}_{point}$ which are discussed in detail in Section \ref{ssec_mesh_cl} and \ref{ssec:point_cl}. In Section \ref{ssec_unsupervised}, we present our unsupervised pipeline with the self- and cross-consistency losses by \cite{zhou2020unsupervised}. However, directly using these losses leads to suboptimal results due to 3D-CoreNet not having a disentangled latent space.
Hence, given an input mesh $\mathbf{x}$, we further separate its latent representation $F(\mathbf{x})\in\mathbb{R}^{N\times D}$ into identity $F_{id}(\mathbf{x})\in\mathbb{R}^{N\times D_{id}}$ and pose $F_{pose}(\mathbf{x})\in\mathbb{R}^{N\times D_{pose}}$ channels. Here, $F(\cdot)$ is the feature extractor, and $F_{id}$ and $F_{pose}$ are the components of $F$ corresponding to the pose and identity channels, and $D_{id} + D_{pose} = D$. Furthermore, we feed \emph{only} the identity channels $F_{id}(\mathbf{x}^{B2})$ to the refinement module as input, but \emph{both} identity and pose channels to the correspondence module as input.

\subsection{Unsupervised pose transfer}
\label{ssec_unsupervised}
\begin{figure*}[t]
\begin{center}
\includegraphics[width=0.99\linewidth]{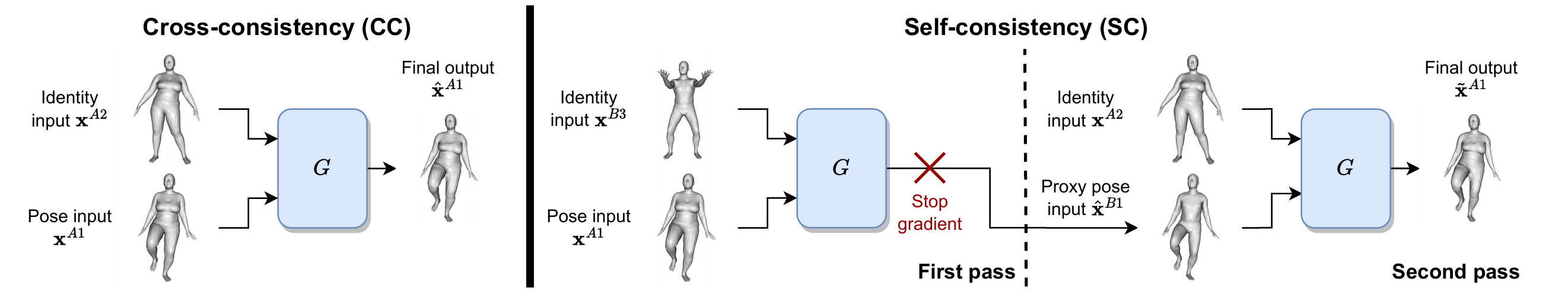}
\end{center}
   \caption{\textbf{Our unsupervised learning pipeline.} During CC, the model receives two inputs with the same identity. During SC, the pose input in the second pass is of a different identity but in the same pose as that in CC, generated by the model itself in the first pass. The model should learn to produce the same output in both stages. This framework is based on \cite{zhou2020unsupervised}.}
\label{fig:uspd}
\end{figure*}

It is clear that training 3D-CoreNet requires the ground truth output with the desired pose and identity. This requires training samples with: (i) the same identity in different poses, \emph{and} (ii) different identities in the same pose. Whilst (i) is easily satisfied, (ii) is more difficult in practice. For instance, if the two inputs come from separate datasets with different sets of identities and poses, there would be no ground truth for the reconstruction loss (Equation \ref{eqn:rec_loss}).

Unsupervised pose transfer with only condition (i) was shown to be possible on registered meshes by \cite{zhou2020unsupervised}, whose main idea is that the network should arrive at the same output in two sub-tasks: (a) when both inputs share a common identity, and (b) when the pose input in (a) is replaced by a different identity with the same pose. Task (a) is called ``cross-consistency'' and is readily available from the dataset. The pose input in task (b) is unavailable but can be generated by the network itself, i.e. ``self-consistency''. However, their GCN model and ARAP deformation require the vertices of both input meshes to be pre-aligned.

Despite these differences, we propose incorporating the cross- and self- consistency losses from \cite{zhou2020unsupervised} into the task of pose transfer on unaligned meshes to enable unsupervised training (see Figure \ref{fig:uspd}). Following \cite{zhou2020unsupervised}, meshes $\mathbf{x}^{A1},\mathbf{x}^{A2}\in\mathbb{R}^{N_{pose}\times 3}$ and $\mathbf{x}^{B3}\in\mathbb{R}^{N_{id}\times 3}$ are required as inputs during training. In addition, whilst vertex order can vary across identities, it must be the same between $\mathbf{x}^{A1}$ and $\mathbf{x}^{A2}$.


\textbf{Cross-consistency (CC).} Given pose input $\mathbf{x}^{A1}$ and identity input $\mathbf{x}^{A2}$, which have the same identity but different poses, the network should reconstruct $\mathbf{x}^{A1}$ through $\hat{\mathbf{x}}^{A1} = G(\mathbf{x}^{A1},\mathbf{x}^{A2})$ (Figure \ref{fig:uspd}, left). This is enforced via
\begin{equation}\label{eqn:cc}
     \mathcal{L}_{cc} = \lambda_{rec}\mathcal{L}_{rec}(\hat{\mathbf{x}}^{A1}; \mathbf{x}^{A1}) + \lambda_{edge}\mathcal{L}_{edge}(\hat{\mathbf{x}}^{A1}; \mathbf{x}^{A2}).
\end{equation}

\textbf{Self-consistency (SC).} CC alone is insufficient as it does not train the network to perform transfers between meshes with different identities. As mentioned previously, the model should reconstruct the same output as that in CC when its pose input is replaced by a different identity with the same pose -- which is usually not available from the training data. \cite{zhou2020unsupervised} proposed to let the network itself generate such examples as proxy inputs for training in a two-pass manner (Figure \ref{fig:uspd}, centre and right). In the first pass, given pose input $\mathbf{x}^{A1}$ and identity input $\mathbf{x}^{B3}$, the network generates a proxy $\hat{\mathbf{x}}^{B1} = G(\mathbf{x}^{A1},\mathbf{x}^{B3})$. This is then used as the pose input in the second pass to reconstruct the initial pose input $\tilde{\mathbf{x}}^{A1} = G(SG(\hat{\mathbf{x}}^{B1}), \mathbf{x}^{A2})$, where $SG$ stops the gradient from passing through. The purpose of $SG$ is to prevent the model from exploiting shortcuts such as using the input $\mathbf{x}^{A1}$ from the first pass to directly reconstruct the output in the second pass. Incidentally, this also avoids extra computational overhead. The SC loss is given by
\begin{equation}\label{eqn:sc}
     \mathcal{L}_{sc} = \lambda_{rec}\mathcal{L}_{rec}(\tilde{\mathbf{x}}^{A1}; \mathbf{x}^{A1}) + \lambda_{edge}\mathcal{L}_{edge}(\tilde{\mathbf{x}}^{A1}; \mathbf{x}^{A2}).
\end{equation}
Finally, the overall unsupervised loss is given by
\begin{equation}
    \mathcal{L}_{us} = \mathcal{L}_{cc} + \mathcal{L}_{sc}.
\end{equation}

\subsection{Mesh contrastive learning}
\label{ssec_mesh_cl}
\begin{figure*}[t]
\begin{center}
\includegraphics[width=0.99\linewidth]{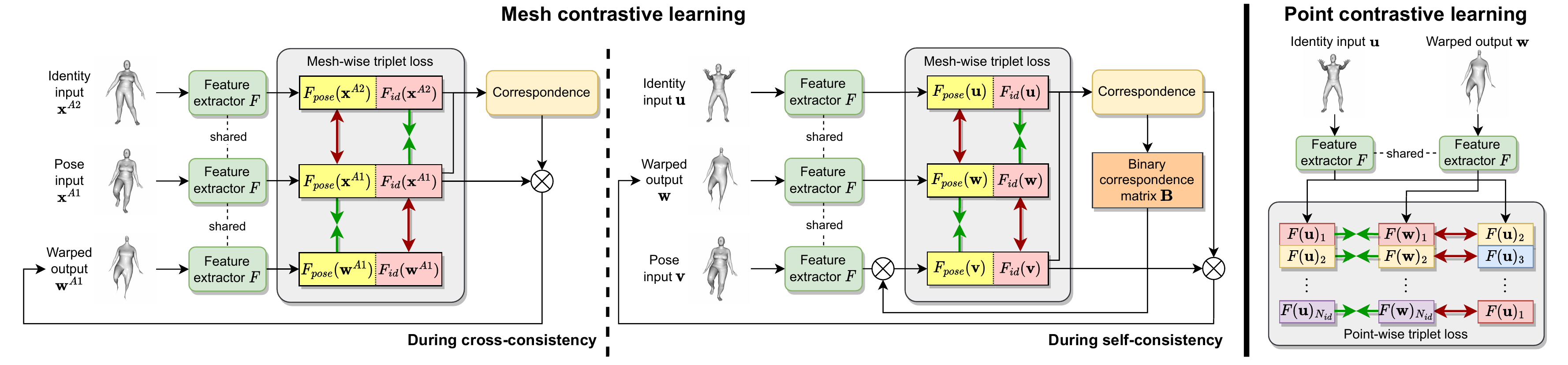}
\end{center}
   \caption{\textbf{Mesh and point contrastive learning.} Left: the triplet formulation for CC, where $\mathbf{x}^{A1}$ and $\mathbf{x}^{A2}$ have the same identity and vertex order but different poses. Centre: the triplet formulation for SC (and supervised pipeline), where $\mathbf{u}$ and $\mathbf{v}$ do not have the same identity, poses, or vertex order. Right: the triplet formulation for point contrastive learning, where $\mathbf{u}$ and $\mathbf{w}$ share the same vertex order.}
\label{fig:disentangle}
\end{figure*}


As mentioned in \ref{ssec:lat_dis}, we disentangle the latents into pose and identity channels. This allows us to impose direct constraints on the meaning of these channels to improve the accuracy of the model output. For instance, we can compare the output against the inputs in terms of pose and identity and impose losses to enforce consistency. In addition, during \emph{unsupervised} learning, the network may also exploit potential shortcuts such as taking both pose and identity information from one input only and ignoring the other input. Disentangling the latent space and imposing additional constraints makes these shortcuts more difficult to exploit.

For these purposes, we propose mesh-level contrastive learning losses for pose and identity. As we cannot compare pose and identity directly in the mesh space, we take a self-supervised approach by feeding meshes through the feature extractor $F$ and imposing the triplet loss \cite{Schroff2015FaceNet} on the latent representations (see Figure \ref{fig:disentangle}). Specifically, given an anchor latent $\mathbf{a}$, a positive latent $\mathbf{p}$, and a negative latent $\mathbf{n}$ which are all in $\mathbb{R}^{N\times D}$, our mesh triplet loss is given by
\begin{equation} \label{eqn:mesh_triplet}
    l(\mathbf{a}, \mathbf{p}, \mathbf{n}) = \left(m + \frac{1}{N}\sum_{j=1}^N d(\mathbf{a}_j, \mathbf{p}_j, \mathbf{n}_j) \right)^{+},
\end{equation}
where $(\cdot)^{+}=\max(0,\cdot)$, $m$ is the margin, and
\begin{equation} \label{eqn:triplet}
    d(\mathbf{a}_j, \mathbf{p}_j, \mathbf{n}_j) = \|\mathbf{a}_j-\mathbf{p}_j\|_2 - \|\mathbf{a}_j-\mathbf{n}_j\|_2,
\end{equation}
where $\mathbf{a}_j,\mathbf{p}_j,\mathbf{n}_j\in\mathbb{R}^D$ are the latents of the $j$-th vertex from $\mathbf{a}, \mathbf{p}, \mathbf{n}$ respectively. In equation \ref{eqn:mesh_triplet}, we enforce the margin $m$ on the whole mesh rather than on individual points since pose and identity are global patterns. We will now discuss how equation \ref{eqn:mesh_triplet} is incorporated into our unsupervised and supervised pipelines.


\textbf{Contrastive learning in CC.} Recall that in CC, the network tries to predict the output from two inputs with the same identity. Given identity input $\mathbf{x}^{A2}$ and pose input $\mathbf{x}^{A1}$, let $\mathbf{w}^{A1}$ be the resulting warped output.
By intuition, the pose representation of $\mathbf{x}^{A1}$ should be closer to that of $\mathbf{w}^{A1}$ than $\mathbf{x}^{A2}$, as the pose of $\mathbf{x}^{A1}$ should be inherited by $\mathbf{w}^{A1}$. Whilst $\mathbf{w}^{A1}$, $\mathbf{x}^{A2}$, and $\mathbf{x}^{A1}$ should all have the same identity, the identity representation of $\mathbf{x}^{A1}$ should be closer to that of $\mathbf{x}^{A2}$ than $\mathbf{w}^{A1}$ as $\mathbf{w}^{A1}$ should generally avoid inheriting the identity representation from the pose input.
Hence, the triplet loss formulation for CC is
\begin{align}\label{eqn:mesh_triplet_cc}
    \begin{split}
        \mathcal{L}_{mesh}^{cc} = & l\left(F_{pose}(\mathbf{x}^{A1}),F_{pose}(\mathbf{w}^{A1}), F_{pose}(\mathbf{x}^{A2})\right) \\
        & + l\left(F_{id}(\mathbf{x}^{A1}),F_{id}(\mathbf{x}^{A2}),F_{id}(\mathbf{w}^{A1})\right)
    \end{split}
\end{align}

We use the warped output here (and later in SC) instead of the final output for two reasons. First, this forces the warped output to be closer to the desired final output, making it easier to be refined. Second, this has a lower computational cost as it does not involve the refinement module.

\textbf{Contrastive learning in SC.} Recall that in SC, the model should produce the same output as that in CC through two consecutive passes, and the two inputs for each pass have different identities and poses. Given identity input $\mathbf{u}\in\mathbb{R}^{N_{id}\times 3}$ and pose input $\mathbf{v}\in\mathbb{R}^{N_{pose}\times 3}$, let $\mathbf{w}\in\mathbb{R}^{N_{id}\times 3}$ be the resulting warped output. Naturally, the pose representation of $\mathbf{w}$ should be closer to that of $\mathbf{v}$ than $\mathbf{u}$, and the identity representation of $\mathbf{w}$ should be closer to that of $\mathbf{u}$ than $\mathbf{v}$. This logic applies to \emph{both} passes within SC.

However, unlike CC, the vertex orders of $\mathbf{u}$ and $\mathbf{v}$ in SC are not aligned. As a result, equations \ref{eqn:mesh_triplet} and \ref{eqn:triplet} cannot be applied directly as they only compare distances between \emph{aligned} vertices. We again take a self-supervised approach to ``reorder'' the feature of $\mathbf{v}$ by utilising the OT matrix $\mathbf{T}$. As the cost matrix for OT is based on the pairwise similarities between point features of $\mathbf{u}$ and $\mathbf{v}$, most entries in $\mathbf{T}$ are made close to zero except for a small portion which are more likely to be corresponding points. Therefore, we use the following binary version of $\mathbf{T}$ to select vertices from $\mathbf{v}$
\begin{equation}\label{eqn:binary_mat}
    \mathbf{B}_{jk} = I\{\mathbf{T}_{jk} = \max_{l}\mathbf{T}_{jl}\},
\end{equation}
where $I$ is the indicator function. In other words, each row of $\mathbf{B}\in\{0,1\}^{N_{id}\times N_{pose}}$ is a binary vector marking the location of the maximum value in the corresponding row of $\mathbf{T}$.
We further constrain the rows of $\mathbf{B}$ to be one-hot in case of multiple maximum entries.
Now, the triplet loss for SC with the ``reordered" pose feature is given by
\begin{align}\label{eqn:mesh_triplet_sc}
    \begin{split}
        \mathcal{L}_{mesh}^{ss} = & l\left(F_{pose}(\mathbf{w}),\mathbf{B}F_{pose}(\mathbf{v}),F_{pose}(\mathbf{u})\right) \\
        & + l\left(F_{id}(\mathbf{w}),F_{id}(\mathbf{u}),\mathbf{B}F_{id}(\mathbf{v})\right).
    \end{split}
\end{align}

  
\textbf{Contrastive learning in supervised pipeline.} Equation \ref{eqn:mesh_triplet_sc} can also be applied to our supervised learning pipeline by replacing $\mathbf{u},\mathbf{v},\mathbf{w}$ with $\mathbf{x}^{B2},\mathbf{x}^{A1},\mathbf{w}^{B1}$, respectively.

\subsection{Point contrastive learning}\label{ssec:point_cl}
Intuitively, the final output would be more accurate if the warped output more closely resembles the ground truth. However, as later experiments will demonstrate, both 3D-CoreNet and $\mathcal{L}_{mesh}$ have a shrinking effect on the warped output, particularly on the head and lower limbs.
We propose to address this problem by enforcing similarity between corresponding points and dissimilarity between non-corresponding points across different meshes.
Specifically, given identity input $\mathbf{u}$ and warped output $\mathbf{w}$, we propose the following triplet loss for point-level contrastive learning
\begin{equation}\label{eqn:point_triplet}
    \mathcal{L}_{point} = \frac{1}{N_{id}}\sum_{j=1}^{N_{id}}\left(m + d(F(\mathbf{w})_j, F(\mathbf{u})_j, F(\mathbf{u})_{k}) \right)^{+},
\end{equation}
where $d$ is from equation \ref{eqn:triplet} and $k\in\{1,\dots,N_{id}\}\setminus\{j\}$. Unlike equation \ref{eqn:mesh_triplet}, the margin $m$ here is enforced on individual points instead of the whole mesh. In our implementation, we set $k=j+1$, and consequently set $F(\mathbf{u})_{N_{id}+1} := F(\mathbf{u})_1$. In other words, the negative points are simply $\mathbf{u}$ with its first point moved to the end and all others shifted down in index by 1. As all input points of a mesh are randomly re-ordered during pre-processing (see Section \ref{sec:experiments}), the negative point can come from any region of the feature $F(\mathbf{u})$ throughout training. In addition, this loss can be applied in all cases including supervised learning and both CC and SC in unsupervised learning.
  

\begin{table*}[t]
    \centering
    \begin{tabularx}{0.99\linewidth}{c|c|X|r|r|r}
        \thickhline 
        \emph{Test set} & \emph{Mode} & \emph{Method} & \textit{PMD $\downarrow$} & \textit{CD $\downarrow$} & \textit{EMD $\downarrow$} \\ \hline
        \multirow{10}{*}{SMPL} & \multirow{6}{*}{S} & (\emph{A}) DT \cite{Sumner2004Deformation} & 1.50 & 3.50 & 22.10 \\ 
        & & (\emph{B}) NPT \cite{Wang2020Neural} & 6.60 & 14.20 & 42.20 \\ 
        & & (\emph{C}) 3D-CoreNet (Baseline) \cite{song20213d}  & 0.36 & 1.18 & 1.35 \\ 
        & & (\emph{D}) {\modelname} (Ours) & \textbf{0.30} & \textbf{1.04} & \textbf{1.15} \\ \cline{3-6} 
        & & (\emph{E}) 3D-CoreNet (Baseline) \cite{song20213d}, 50\% labelled & 4.99 & 10.34 & 10.13 \\ 
        & & (\emph{F}) {\modelname} (Ours), 50\% labelled & \textbf{3.69} & \textbf{6.67} & \textbf{9.25} \\ \cline{2-6} 
        & SS & (\emph{G}) {\modelname} (Ours), 50\% labelled + 50\% unlabelled & \textbf{0.40} & \textbf{1.32} & \textbf{1.45} \\ \cline{2-6} 
        & \multirow{3}{*}{US} & (\emph{H}) 3D-CoreNet (Baseline) \cite{song20213d}, without LD & 14.09 & 32.34 & 18.85 \\ 
        & & (\emph{I}) 3D-CoreNet (Baseline) \cite{song20213d}, with LD & 0.76 & 2.20 & 2.58 \\ 
        & & (\emph{J}) {\modelname} (Ours) & \textbf{0.56} & \textbf{1.71} & \textbf{1.83} \\ \hline 
        \multirow{8}{*}{SMAL} & \multirow{6}{*}{S} & (\emph{K}) DT \cite{Sumner2004Deformation} & 133.70 & 357.70 & 159.00 \\
        & & (\emph{L}) NPT \cite{Wang2020Neural} & 67.50 & 145.20 & 116.50 \\
        & & (\emph{M}) 3D-CoreNet (Baseline) \cite{song20213d} & 27.04 & 51.55 & 29.31 \\ 
        & & (\emph{N}) {\modelname} (Ours) & \textbf{25.34} & \textbf{47.81} & \textbf{26.22} \\ \cline{3-6}
        & & (\emph{O}) 3D-CoreNet (Baseline) \cite{song20213d}, 50\% labelled & 59.66 & 130.99 & 49.26 \\
        & & (\emph{P}) {\modelname} (Ours), 50\% labelled & \textbf{51.36} & \textbf{112.25} & \textbf{43.80} \\ \cline{2-6}
        & SS & (\emph{Q}) {\modelname} (Ours), 50\% labelled + 50\% unlabelled & \textbf{29.80} & \textbf{55.16} & \textbf{30.85} \\ \cline{2-6}
        & US & (\emph{R}) {\modelname} (Ours) & \textbf{29.29} & \textbf{54.64} & \textbf{30.00} \\ \hline
        \multirow{6}{*}{DFAUST} & \multirow{3}{*}{US} & (\emph{S}) 3D-CoreNet (Baseline) \cite{song20213d}, without LD & 468.10 & 441.74 & 128.08 \\
        & & (\emph{T}) 3D-CoreNet (Baseline) \cite{song20213d}, with LD & 92.45 & 176.21 & 55.71 \\
        & & (\emph{U}) {\modelname} (Ours) & \textbf{61.22} & \textbf{60.97} & \textbf{32.07} \\ \cline{2-6}
        & \multirow{2}{*}{S} & (\emph{C}) 3D-CoreNet (Baseline) \cite{song20213d}, trained on SMPL only & 138.56 & 159.11 & 58.95 \\
        & & (\emph{D}) {\modelname} (Ours), trained on SMPL only & \textbf{120.10} & \textbf{133.09} & \textbf{49.75} \\ \cline{2-6}
        & SS & (\emph{V}) {\modelname} (Ours), SMPL labelled + DFAUST unlabelled & \textbf{36.23} & \textbf{35.69} & \textbf{23.30} \\
        \thickhline
    \end{tabularx}
    \caption{\textbf{Quantitative results.} PMD and CD are in units of $10^{-4}$, and EMD is in units of $10^{-3}$. Lower values are better. The modes ``S", ``SS", and ``US" are short for ``supervised", ``semi-supervised", and ``unsupervised", respectively. ``LD" is short for ``latent disentanglement" which is described in Section \ref{ssec:lat_dis}. \emph{Note: Training and test sets are from the same dataset unless otherwise specified}.}
    \label{tab:quantitative_3d}
\end{table*}

\subsection{Overall training losses}
The overall objective of our framework when ground truth is available is given by the labelled loss:
\begin{equation} \label{eqn:labelled}
     \mathcal{L}_L = \mathcal{L}_{s} + \lambda_{m,s} \mathcal{L}_{mesh}^{ss} + \lambda_p \mathcal{L}_{point}.
\end{equation}
When ground truth is unavailable, we instead minimise:
\begin{equation} \label{eqn:unlabelled}
    \mathcal{L}_U = \mathcal{L}_{us} + \lambda_{m,c} \mathcal{L}_{mesh}^{cc} + \lambda_{m,s} \mathcal{L}_{mesh}^{ss} + \lambda_p\mathcal{L}_{point}.
\end{equation}
In our full models, we set $\lambda_{m,s} =\lambda_{m,c} = \lambda_p = 1$.


\section{Experiments}
\label{sec:experiments}

\subsection{Data}
\textbf{SMPL.} This is a synthetic dataset \cite{Wang2020Neural} of 24,000 human meshes generated by SMPL \cite{Loper2015SMPL} from 30 identities and 800 poses. Each mesh has 6,890 vertices. Ground truths are available for evaluation and supervised learning as all identities share a common set of poses. Following \cite{song20213d}, we randomly sample a training set of 4,000 meshes from a randomly sampled list of 16 identities and 400 poses. For evaluation, we use the same fixed set of 400 mesh pairs as \cite{song20213d} which are randomly sampled from the remaining 14 identities and 200 poses that are unseen during training.

\textbf{SMAL.} This is a synthetic dataset of 24,600 animal meshes generated by SMAL \cite{Zuffi2017SMAL} from 41 identities and 600 poses. Each mesh has 3,889 vertices. All identities also share a common set of poses. Following \cite{song20213d}, we randomly select a training pool of 11,600 meshes from a random list of 29 identities and 400 poses. For evaluation, we again use the same fixed set of 400 mesh pairs as \cite{song20213d} from the unseen 12 identities and 200 poses. Compared to SMPL, this is a more challenging dataset as it consists of a wider variety of shapes and sizes of animals.

\textbf{DFAUST.} Unlike SMPL and SMAL, which are synthetic datasets, DFAUST \cite{Bogo2017DFAUST} is a more challenging collection of registered human meshes obtained from real 3D scans which allows us to validate our model in realistic unsupervised settings. The dataset consists of 10 subjects, 5 males and 5 females, each performing multiple motion sequences. There are no direct ground truths available since different subjects are not in precisely the same pose, and as a result we use the SMPL+H model \cite{MANO2017Romero} to generate pseudo ground truths for evaluation. We randomly select 4,000 meshes from 3 males and 3 females for training and 399 meshes from the unseen subjects for evaluation. 

\textbf{MG.} The Multi-Garment (MG) dataset \cite{bhatnagar2019mgn} includes registered meshes of humans in clothing from real 3D scans. Each mesh has 27,554 vertices -- significantly more than SMPL and DFAUST. We use MG to qualitatively validate our method's ability to handle complex unseen topologies.

\begin{figure*}[t]
\begin{center}
\includegraphics[width=0.99\linewidth]{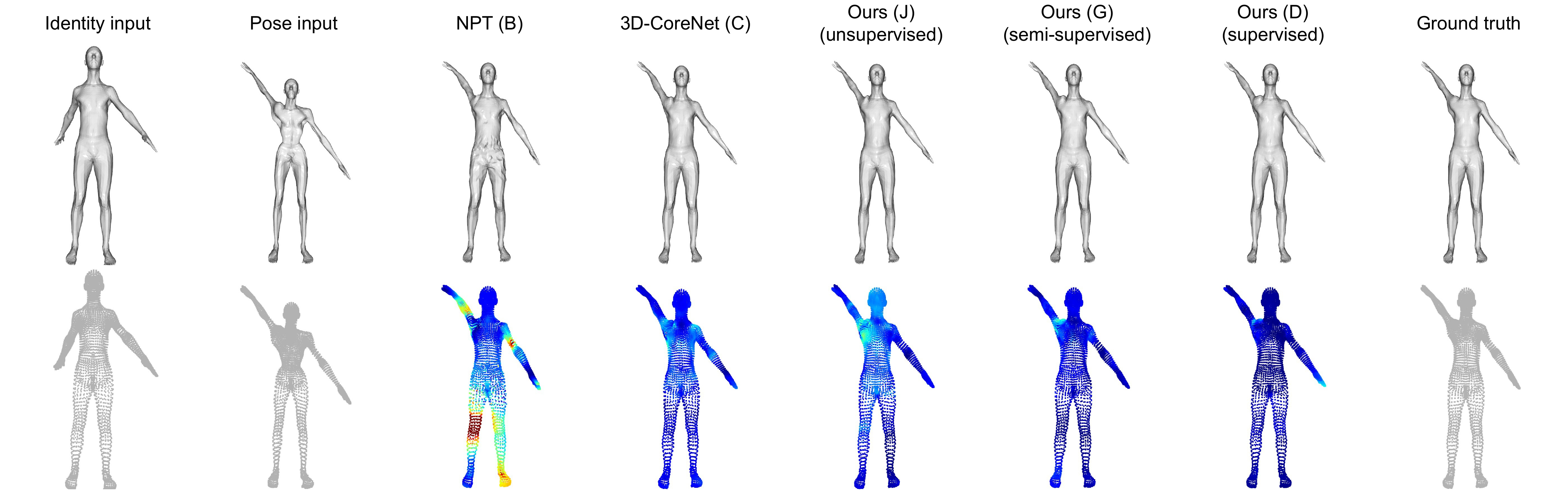}
\end{center}
   \caption{\textbf{Qualitative comparison on unseen SMPL inputs.} This shows pose transfer results using prior and our methods (labels defined in Table \ref{tab:quantitative_3d}) trained on SMPL. The first and second rows show the rendered surfaces and point clouds respectively. The colours of the model output point clouds represent PMDs against the ground truth, with dark red and dark blue indicating high and low PMDs respectively.}
\label{fig:qualitative_3d}
\end{figure*}

\begin{figure*}[t]
\begin{center}
\includegraphics[width=0.99\linewidth]{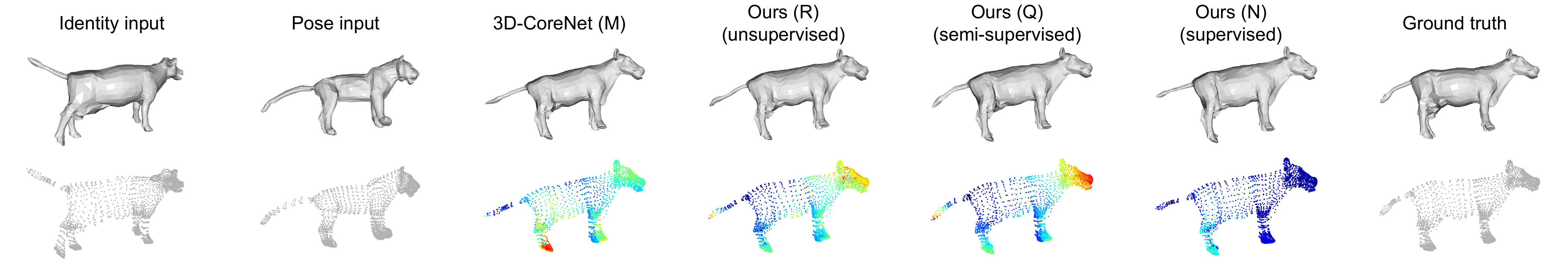}
\end{center}
   \caption{\textbf{Qualitative comparison on unseen SMAL inputs.} Similar to Figure \ref{fig:qualitative_3d} (with method labels defined in Table \ref{tab:quantitative_3d}), the first and second rows are the rendered surfaces and point clouds respectively, with PMD heatmaps visualised on model output point clouds.}
\label{fig:qualitative_animal}
\end{figure*}

\begin{figure*}[t]
\begin{center}
\includegraphics[width=0.99\linewidth]{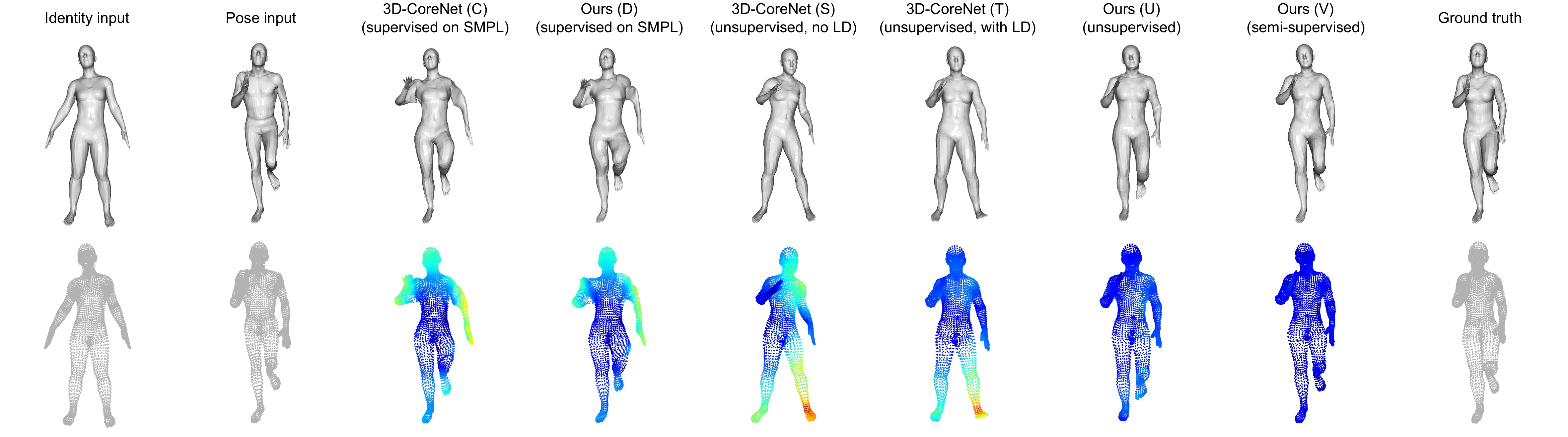}
\end{center}
   \caption{\textbf{Qualitative comparison on unseen DFAUST inputs.} Similar to Figure \ref{fig:qualitative_3d} (with method labels defined in Table \ref{tab:quantitative_3d}), the first and second rows are the rendered surfaces and point clouds respectively, with PMD heatmaps visualised on model output point clouds.}
\label{fig:qualitative_dfaust}
\end{figure*}

\subsection{Implementation details} \label{ssec:implementation}
\textbf{Pre-processing.} We pre-process all inputs in two steps. First, the vertices of each pair of pose and identity inputs are \emph{randomly and independently re-ordered} to remove the correspondence between them. Second, they are zero-centred based on their bounding boxes. We perform the same procedure for \emph{both} training and evaluation.

\textbf{Parameter settings.} Our experiments are based on the official implementation of 3D-CoreNet \cite{song20213d}.
We set a batch size of 2 for all experiments. Following \cite{song20213d}, all hyperparameters are kept at their default values including $\lambda_{rec}=1000$ and $\lambda_{edge}=0.5$. As for $\mathcal{L}_{mesh}$ and $\mathcal{L}_{point}$, we set the margin at $m=1$. The network is trained for 200 epochs using the Adam optimiser \cite{Kingma2015Adam} with an initial learning rate of $1\times10^{-4}$ which is kept constant for the first 100 epochs and then linearly decayed to 0 in the last 100 epochs. 


\textbf{Training.} The detailed training procedures for supervised, unsupervised, and semi-supervised settings are in Algorithms 1, 2, and 3 in the Appendix, respectively. For semi-supervised learning in SMPL or SMAL (Table \ref{tab:quantitative_3d} methods (\emph{G}) and (\emph{Q})), we alternate iterations between optimising $\mathcal{L}_{L}$ and $\mathcal{L}_{U}$. On the other hand, for semi-supervised learning on SMPL and DFAUST (Table \ref{tab:quantitative_3d} method (\emph{V})), as we would like to run inference on DFAUST with the final model, we train the first 100 epochs on SMPL in a supervised manner and the remaining epochs on DFAUST in an unsupervised manner. During an unlabelled iteration, we also allow either pose or identity input (not both) to come from the labelled dataset (see Algorithm 3 in the Appendix).

\subsection{Evaluation}
We evaluate model performance at epoch 200 by comparing its outputs against ground truths using three metrics.

\textbf{Pointwise Mesh Distance (PMD).} The PMD \cite{Wang2020Neural,song20213d} is in the same form as Equation \ref{eqn:rec_loss} and takes into account both the positions and ordering of the vertices measured. 

\textbf{Chamfer Distance (CD).} The CD \cite{Fan2017CD} measures the discrepancy between two point clouds without taking into account the ordering of their points. Given point clouds $P,Q\subset\mathbb{R}^3$, the CD between $P$ and $D$ is given by
\begin{equation}
    \frac{1}{|P|}\sum_{\mathbf{p}\in P} \min_{\mathbf{q}\in Q}\|\mathbf{p} - \mathbf{q} \|_2^2 + \frac{1}{|Q|}\sum_{\mathbf{q}\in Q} \min_{\mathbf{p}\in P}\|\mathbf{p} - \mathbf{q} \|_2^2.
\end{equation}

\textbf{Earth Mover's Distance (EMD).} The EMD \cite{Fan2017CD} first solves the assignment problem between two point clouds before taking the pointwise distance by aligning the two point clouds using the resulting assignment function. Specifically, given point clouds $P,Q\subset\mathbb{R}^3$ such that $|P|=|Q|$, the EMD between $P$ and $Q$ is given by
\begin{equation}
    \min_{\phi:P\rightarrow Q} \frac{1}{|P|} \sum_{\mathbf{p}\in P} \| \mathbf{p} - \phi(\mathbf{p}) \|_2,
\end{equation}
where $\phi$ is a bijective mapping from $P$ to $Q$.

For all metrics, lower values are better. As mentioned in section \ref{ssec:implementation}, we randomly and independently re-order the input vertices in evaluations (similar to training) but with a \emph{fixed} seed for all experiments to ensure fairness.


\subsection{Quantitative results}

We compare our method with previous state-of-the-art pose transfer models using the aforementioned metrics and evaluation procedure (see Table \ref{tab:quantitative_3d}). For DT \cite{Sumner2004Deformation} and NPT \cite{Wang2020Neural}, we quote the results obtained by \cite{song20213d}. For 3D-CoreNet \cite{song20213d}, we retrain their model from scratch using their official implementation. As we use a smaller batch size during training than that used in the original 3D-CoreNet work \cite{song20213d} but keep the total number of epochs unchanged, there are more gradient descent steps during our training which causes the baseline numbers we obtained to be different from the original numbers reported by \cite{song20213d}. However, our replicated results are close to the original ones or even better. We also emphasise that we keep all data pre-processing, training, and evaluation procedures identical across experiments to ensure any comparisons between the baseline and our method are fair.

We outperform prior state-of-the-art methods in the supervised setting on SMPL and SMAL as shown by method (\emph{D}) and (\emph{N}) in Table \ref{tab:quantitative_3d}. Our method also enables unsupervised learning in (\emph{J}) and (\emph{R}) where we achieve results comparable to the supervised ones while still outperforming earlier supervised methods DT and NPT. Training 3D-CoreNet \cite{song20213d} directly using the CC and SC losses for unsupervised learning \cite{zhou2020unsupervised} does not yield ideal results as shown by (\emph{H}), but applying our latent space disentanglement leads to a substantial improvement as shown by (\emph{I}) and (\emph{J}). Similar improvement can also be observed on the more challenging DFAUST dataset as shown by (\emph{S}), (\emph{T}), and (\emph{U}). This demonstrates that it can be difficult for the model to disentangle pose and identity effectively in an unsupervised setting without proper architectural constraints.

Table \ref{tab:quantitative_3d} also demonstrates that our method is more effective in settings with limited labelled data. In methods (\emph{E}), (\emph{F}), (\emph{O}), and (\emph{P}), we randomly remove $50\%$ identities and $50\%$ poses from the training set, which means the model only sees $25\%$ of all available meshes during training. As expected, this leads to a dramatic increase in PMD and CD which indicates reduced model generalisability. However, our model is still able to outperform 3D-CoreNet under this limitation. Since our method also supports unsupervised learning, we can introduce meshes with the remaining $50\%$ identities and $50\%$ poses to our model as unlabelled samples in addition to the labelled ones. Note that the model still only sees $50\%$ of all available training meshes since meshes whose identity is in the labelled set and whose pose is in the unlabelled set (and vice versa) are not included. Under this semi-supervised setting, our model achieves substantial improvement in accuracy as shown by (\emph{G}) and (\emph{Q}), approaching the fully supervised results. In addition, by using all SMPL training data as the labelled set and all DFAUST training data as the unlabelled set, we achieve further improvement when testing on DFAUST as shown by (\emph{V}). 


\subsection{Qualitative results}

\begin{figure*}[t]
\begin{center}
\includegraphics[width=0.99\linewidth]{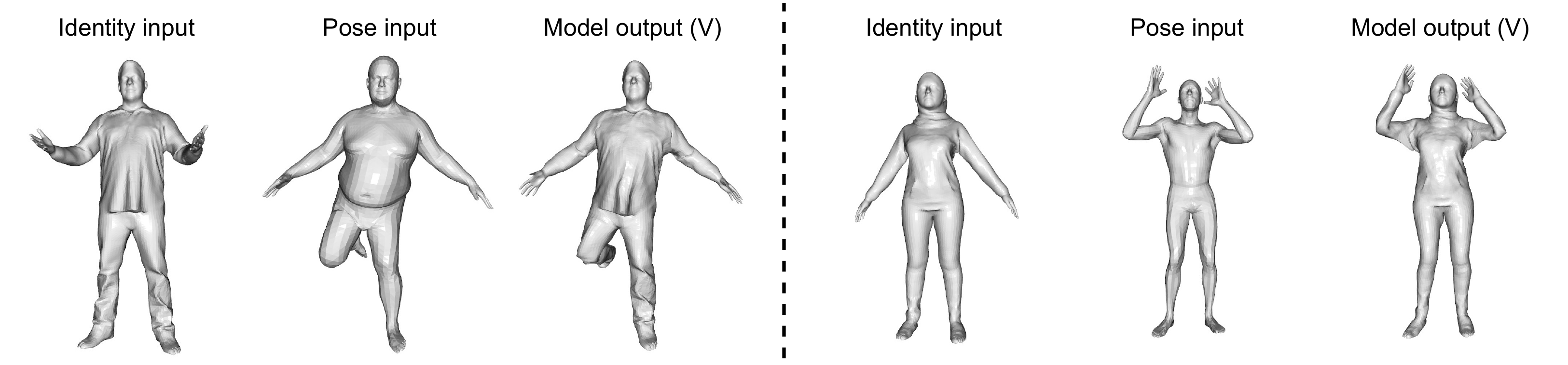}
\end{center}
   \caption{\textbf{Qualitative results on unseen MG and DFAUST inputs.} The above shows two instances of pose transfer using our semi-supervised model (\emph{V}) defined in Table \ref{tab:quantitative_3d}, with MG meshes as identity inputs and DFAUST meshes as pose inputs.}
\label{fig:qualitative_mg}
\end{figure*}

\begin{figure*}[t]
\begin{center}
\includegraphics[width=0.99\linewidth]{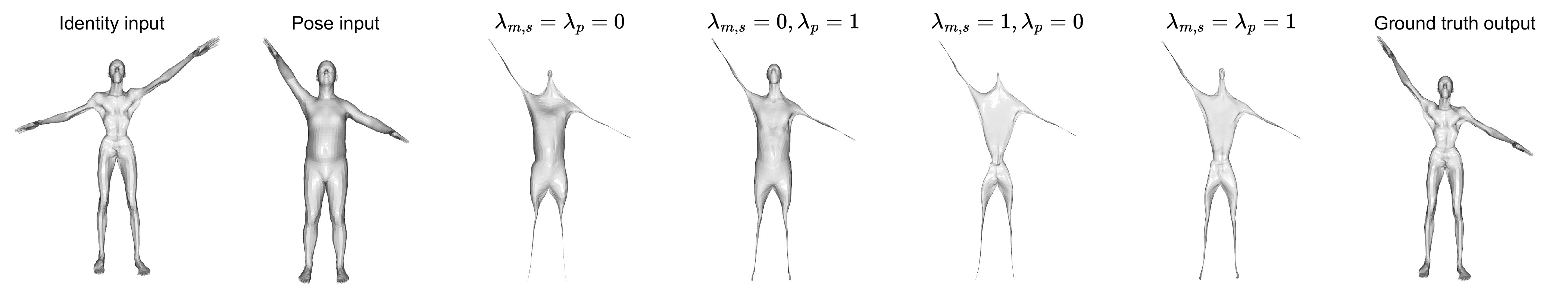}
\end{center}
   \caption{\textbf{Warped outputs in the ablation study (supervised).} This is a visual comparison of the warped outputs from the supervised models in the ablation study (Table \ref{tab:ablation}), with the input and ground truth output meshes as references.}
\label{fig:ablation_3d}
\end{figure*}


We visually compare the outputs of various methods on SMPL and SMAL in Figures \ref{fig:qualitative_3d} and \ref{fig:qualitative_animal}, respectively. We can observe that the output surfaces of our method are smoother compared to NPT which produces numerous concavities. We visualise the PMDs between the outputs and ground truths through heatmaps which show that our method generates more accurate outputs indicated by the darker blue. We also visualise various models on DFAUST in Figure \ref{fig:qualitative_dfaust}, which clearly shows that our unsupervised model (\emph{U}) is more accurate than (\emph{S}) and (\emph{T}) which do not employ our contrastive losses, and our semi-supervised model (\emph{V}) trained on both SMPL and DFAUST outperforms the supervised models (\emph{C}) and (\emph{D}) trained only on SMPL. Finally, Figure \ref{fig:qualitative_mg} demonstrates that our model can handle complex unseen input meshes with different topologies and vertex numbers. Additional results can be found in the Appendix.

\subsection{Ablation study}
\begin{table}[t]
    \centering
    \begin{tabularx}{0.99\linewidth}{c|c|c|c|c|c|c}
        \thickhline
        \emph{Mode} & $\lambda_{m,c}$ & $\lambda_{m,s}$ & $\lambda_p$ & \emph{PMD $\downarrow$} & \emph{CD $\downarrow$} & \emph{EMD $\downarrow$} \\ \hline
        \multirow{4}{*}{S} & N/A & 0 & 0 & 0.36 & 1.18 & 1.35 \\
        & N/A & 0 & 1 & 0.35 & 1.16 & 1.26 \\
        & N/A & 1 & 0 & 0.31 & 1.08 & 1.17 \\
        & N/A & 1 & 1 & \textbf{0.30} & \textbf{1.04} & \textbf{1.15} \\ \hline
        \multirow{5}{*}{US} & 0 & 0 & 0 & 0.76 & 2.20 & 2.58 \\
        & 0 & 0 & 1 & 0.59 & 1.81 & 2.00 \\
        & 1 & 0 & 0 & 0.60 & 1.81 & 1.98 \\
        & 1 & 1 & 0 & 0.59 & 1.79 & 1.96 \\
        & 1 & 1 & 1 & \textbf{0.56} & \textbf{1.71} & \textbf{1.83} \\ \thickhline
    \end{tabularx}
    \caption{\textbf{Ablation study.} PMD and CD are in units of $10^{-4}$, and EMD is in units of $10^{-3}$. The modes ``S" and ``US" are short for ``supervised" and ``unsupervised" respectively. }
    \label{tab:ablation}
\end{table}

In Table \ref{tab:ablation}, we study the individual effectiveness of our proposed losses $\mathcal{L}_{mesh}$ and $\mathcal{L}_{point}$ on SMPL by changing their associated weights $\lambda_{m,c}$, $\lambda_{m,s}$, and $\lambda_p$. It can be seen that setting any of them to 1 can improve the result, and setting all to 1 yields the best result. In the unsupervised setting, we also observe that setting $\lambda_{m,c}=\lambda_{m,s}=1$ is more effective than setting $\lambda_{m,c}=1$ only. The warped outputs of the supervised experiments are shown in Figure \ref{fig:ablation_3d}. Setting $\lambda_p=1$ leads to a visible improvement in the quality of the warped output particularly in the head and lower limbs, mitigating the shrinking effect as mentioned in Section \ref{ssec:point_cl}. On the other hand, setting $\lambda_{m,s}=1$ allows the warped output to have a better resemblance with the ground truth. This is shown by the corresponding warped outputs in Figure \ref{fig:ablation_3d} which are thinner compared to the ones with $\lambda_{m,s}=0$. However, this causes a side effect where the head and limbs are drastically shrunk. Combining both losses leads to an improved resemblance between the warped output and ground truth whilst reducing excessive shrinking.


\section{Conclusion}
\label{sec:conclusion}


We proposed {\modelname}, a self-supervised 3D pose transfer framework without requiring correspondence labels or ground truth outputs for supervision. We introduce disentangled latent spaces for pose and identity to improve unsupervised learning, and mesh and point level contrastive learning to improve the model's intermediate output and in turn, final output. We achieve state-of-the-art results in supervised learning, and competitive results in unsupervised and semi-supervised settings that are generalisable to unseen human and animal data with complex topologies.

\section{Acknowledgements}
This work was supported by the Croucher Foundation, Huawei Consumer Business Group, NST grant (CRC 21011, MSIT), and KOCCA grant (R2022020028, MCST).

{\small
\bibliographystyle{ieee_fullname}
\bibliography{egbib}
}

\clearpage
\appendix
\section{Architecture}
The architecture of our {\modelname} is based on that of 3D-CoreNet \cite{song20213d} and is shown in Table \ref{tab:architecture}. We use $F$, $C$, and $R$ to denote the feature extraction, correspondence, and refinement modules, respectively. We let $N_{id}$ and $N_{pose}$ be the number of vertices in the input identity and pose meshes, respectively. One major difference between our method and 3D-CoreNet is that the output of our $F$ module is split into identity and pose parts, each with a size of $N_{id}\times 128$ (for identity input) or $N_{pose}\times 128$ (for pose input), whereas 3D-CoreNet outputs a single feature of size $N_{id}\times 256$ or $N_{pose}\times 256$. In addition, we only feed the identity part of the latent feature to the refinement module, resulting in a slightly lower total number of trainable parameters (23.8 million compared to 24.5 million in 3D-CoreNet).

\begin{table}[h]
    \centering
    \begin{tabularx}{\linewidth}{c|c|Y}
        \thickhline
        \emph{Module} & \emph{Layer} & \emph{Output shape(s)} \\ \hline
        \multirow{5}{*}{$F$} & Conv1D, $1$ & $N_{id}\times 64$, $N_{pose}\times 64$ \\
        & Conv1D, $1$ & $N_{id}\times 128$, $N_{pose}\times 128$ \\
        & Conv1D, $1$ & $N_{id}\times 256$, $N_{pose}\times 256$\\ \cline{2-3}
        & \multirow{2}{*}{RB $\times 4$} & $N_{id}\times 128$, $N_{id}\times 128$ \\
        & & $N_{pose}\times 128$, $N_{pose}\times 128$ \\ \hline
        \multirow{4}{*}{$C$} & Conv1D, $1$ & $N_{id}\times 256$ \\
        & Conv1D, $1$ & $N_{pose}\times 256$ \\
        & OT matrix & $N_{id}\times N_{pose}$ \\
        & Warped & $N_{id}\times 3$ \\ \hline
        \multirow{8}{*}{$R$} & Conv1D, $3$ & $N_{id}\times 1024$ \\
        & Conv1D, $1$ & $N_{id}\times 1024$ \\
        & ElaIN RB & $N_{id}\times 1024$ \\
        & Conv1D, $1$ & $N_{id}\times 512$ \\
        & ElaIN RB & $N_{id}\times 512$ \\
        & Conv1D, $1$ & $N_{id}\times 256$ \\
        & ElaIN RB & $N_{id}\times 256$ \\
        & Conv1D, $1$ & $N_{id}\times 3$ \\ \thickhline
    \end{tabularx}
    \caption{\textbf{The layers of {\modelname} in detail.} We use ``Conv1D, $K$'' denote a 1D  convolutional layer with kernel size $K$, ``RB'' for a residual block, and RB $\times4$ means the same RB architecture repeated 4 times in succession. When a layer has multiple output shapes, that layer is shared between those outputs.}
    \label{tab:architecture}
\end{table}

\begin{algorithm}[t]
\caption{\emph{MAPConNet in Supervised Learning.}}
\label{alg:sup}
	\begin{algorithmic}[1]
	    \REQUIRE Dataset of triplets $\{(\mathbf{x}^{A1}, \mathbf{x}^{B2}, \mathbf{x}^{B1}),\ldots\}$, wherein poses must be aligned across identities.
	    \STATE \textbf{Initialise:} Generator $G$ with trainable weights $\theta_G$, feature extractor $F$ (which is a part of $G$), time step $t=0$, initial learning rate $\eta_0$, maximum time step $t_{max}$.
	    \REPEAT
	    \STATE \label{alg:line:sup_data} Retrieve samples $\mathbf{x}^{A1}\in\mathbb{R}^{N_{pose}\times 3}$ and $\mathbf{x}^{B2}, \mathbf{x}^{B1}\in\mathbb{R}^{N_{id}\times 3}$ whose vertices are randomly re-ordered so that $\mathbf{x}^{B1}$ is aligned with $\mathbf{x}^{B2}$ but \emph{not} $\mathbf{x}^{A1}$.
	    \STATE \label{alg:line:sup_start} With $\mathbf{x}^{A1}$ as the pose input, $\mathbf{x}^{B2}$ as the identity input, compute OT matrix $\mathbf{T}\in\mathbb{R}^{N_{id}\times N_{pose}}$, and $\mathbf{B}$ from $\mathbf{T}$ (Equation \ref{eqn:binary_mat}).
	    \STATE Generate warped output $\mathbf{w}^{B1}=\mathbf{T}\mathbf{x}^{A1}$ and final output $\hat{\mathbf{x}}^{B1}\in\mathbb{R}^{N_{id}\times 3}$.
	    \STATE Compute $\mathcal{L}_{s}$ (Equation \ref{eqn:sup}):
	    \begin{equation}
	        \lambda_{rec}\mathcal{L}_{rec}(\hat{\mathbf{x}}^{B1}; \mathbf{x}^{B1}) + \lambda_{edge}\mathcal{L}_{edge}(\hat{\mathbf{x}}^{B1}; \mathbf{x}^{B2}),
	    \end{equation}
	    where $\mathbf{x}^{B1}$ is the ground truth.
	    \STATE Compute $\mathcal{L}_{mesh}^{ss}$ (Equation \ref{eqn:mesh_triplet_sc}):
	    \begin{align}\label{eqn:mesh_triplet_sc_sup}
            \begin{split}
                 & l\left(F_{pose}(\mathbf{w}^{B1}),\mathbf{B}F_{pose}(\mathbf{x}^{A1}),F_{pose}(\mathbf{x}^{B2})\right) \\
                 + & l\left(F_{id}(\mathbf{w}^{B1}),F_{id}(\mathbf{x}^{B2}),\mathbf{B}F_{id}(\mathbf{x}^{A1})\right).
            \end{split}
        \end{align}
        \STATE Compute $\mathcal{L}_{point}$ (Equation \ref{eqn:point_triplet}):
        \begin{equation}
            \frac{1}{N_{id}}\sum_{j=1}^{N_{id}} (m + d(F(\mathbf{w}^{B1})_j, F(\mathbf{x}^{B2})_j, F(\mathbf{x}^{B2})_{k}) )^{+}.
        \end{equation}
	    \STATE \label{alg:line:sup_end} Compute gradients and update:
            \begin{equation}
                \theta_G \leftarrow \theta_G - \eta_t \frac{\partial \mathcal{L}_{L}}{\partial \theta_G},
            \end{equation}
        where
        \begin{equation}
            \mathcal{L}_{L} = \mathcal{L}_{s} + \mathcal{L}_{mesh}^{ss} + \mathcal{L}_{point}.
        \end{equation}
        \STATE $t\leftarrow t+1$.
	    \UNTIL{$t=t_{max}$.}
	\end{algorithmic}
\end{algorithm}

\begin{algorithm*}
\caption{\emph{MAPConNet in Unsupervised Learning.}}
\label{alg:unsup}
	\begin{algorithmic}[1]
	    \REQUIRE Dataset of triplets $\{(\mathbf{x}^{A1}, \mathbf{x}^{A2}, \mathbf{x}^{B3}),\ldots\}$, wherein poses do not have to align across identities.
	    \STATE \textbf{Initialise:} Generator $G$ with trainable weights $\theta_G$, feature extractor $F$ (which is a part of $G$), time step $t=0$, initial learning rate $\eta_0$, maximum time step $t_{max}$.
	    \REPEAT
	    \STATE \label{alg:line:unsup_data} Retrieve meshes $\mathbf{x}^{A1},\mathbf{x}^{A2}\in\mathbb{R}^{N_{pose}\times 3}$ and $\mathbf{x}^{B3}\in\mathbb{R}^{N_{id}\times 3}$ whose vertices are randomly re-ordered so that $\mathbf{x}^{A1}$ is aligned with $\mathbf{x}^{A2}$ but \emph{not} $\mathbf{x}^{B3}$.
	    \STATE \label{alg:line:unsup_start} With $\mathbf{x}^{A1}$ as pose input, $\mathbf{x}^{A2}$ as identity input, compute OT matrix $\mathbf{T}_{cc}\in\mathbb{R}^{N_{pose}\times N_{pose}}$.
	    \STATE Generate warped output $\mathbf{w}^{A1}=\mathbf{T}_{cc}\mathbf{x}^{A1}$ and final output $\hat{\mathbf{x}}^{A1}\in\mathbb{R}^{N_{pose}\times 3}$.
	    \STATE Compute $\mathcal{L}_{cc}$ (Equation \ref{eqn:cc}):
	    \begin{equation}
              \lambda_{rec}\mathcal{L}_{rec}(\hat{\mathbf{x}}^{A1}; \mathbf{x}^{A1}) + \lambda_{edge}\mathcal{L}_{edge}(\hat{\mathbf{x}}^{A1}; \mathbf{x}^{A2}).
        \end{equation}
        \STATE With $\mathbf{x}^{A1}$ as pose input, $\mathbf{x}^{B3}$ as identity input, compute OT matrix $\mathbf{T}\in\mathbb{R}^{N_{id}\times N_{pose}}$, and $\mathbf{B}$ from $\mathbf{T}$ (Equation \ref{eqn:binary_mat}).
        \STATE Generate warped output $\mathbf{w}^{B1}=\mathbf{T}\mathbf{x}^{A1}$ and final output $\hat{\mathbf{x}}^{B1}\in\mathbb{R}^{N_{id}\times 3}$.
        \STATE With $SG(\hat{\mathbf{x}}^{B1})$ as pose input, $\mathbf{x}^{A2}$ as identity input, where $SG$ stops the backward gradient flow, compute OT matrix $\mathbf{T}'\in\mathbb{R}^{N_{pose}\times N_{id}}$, and $\mathbf{B}'$ from $\mathbf{T}'$ (Equation \ref{eqn:binary_mat}).
        \STATE Generate warped output $\tilde{\mathbf{w}}^{A1}=\mathbf{T}'SG(\hat{\mathbf{x}}^{B1})$ and final output $\tilde{\mathbf{x}}^{A1}\in\mathbb{R}^{N_{pose}\times 3}$.
        \STATE Compute $\mathcal{L}_{sc}$ (Equation \ref{eqn:sc}):
        \begin{equation}
             \lambda_{rec}\mathcal{L}_{rec}(\tilde{\mathbf{x}}^{A1}; \mathbf{x}^{A1}) + \lambda_{edge}\mathcal{L}_{edge}(\tilde{\mathbf{x}}^{A1}; \mathbf{x}^{A2}).
        \end{equation}
	    \STATE Compute $\mathcal{L}_{mesh}^{cc}$ (Equation \ref{eqn:mesh_triplet_cc}):
	    \begin{equation}
            l\left(F_{pose}(\mathbf{x}^{A1}),F_{pose}(\mathbf{w}^{A1}), F_{pose}(\mathbf{x}^{A2})\right) + l\left(F_{id}(\mathbf{x}^{A1}),F_{id}(\mathbf{x}^{A2}),F_{id}(\mathbf{w}^{A1})\right)
        \end{equation}
        \STATE Compute $\mathcal{L}_{mesh}^{ss} = \mathcal{L}_{mesh}^{ss,1} + \mathcal{L}_{mesh}^{ss,2}$ (Equation \ref{eqn:mesh_triplet_sc}), where:
        \begin{align}
             \mathcal{L}_{mesh}^{ss,1} &= l\left(F_{pose}(\mathbf{w}^{B1}),\mathbf{B} F_{pose}(\mathbf{x}^{A1}),F_{pose}(\mathbf{x}^{B3})\right)
             + l\left(F_{id}(\mathbf{w}^{B1}),F_{id}(\mathbf{x}^{B3}),\mathbf{B} F_{id}(\mathbf{x}^{A1})\right), \\
             \mathcal{L}_{mesh}^{ss,2} &= l\left(F_{pose}(\tilde{\mathbf{w}}^{A1}),\mathbf{B}' F_{pose}(SG(\hat{\mathbf{x}}^{B1})),F_{pose}(\mathbf{x}^{A2})\right)
             + l\left(F_{id}(\tilde{\mathbf{w}}^{A1}),F_{id}(\mathbf{x}^{A2}),\mathbf{B}' F_{id}(SG(\hat{\mathbf{x}}^{B1}))\right).
        \end{align}
        \STATE Compute $\mathcal{L}_{point} = \mathcal{L}_{point}^{cc} + \mathcal{L}_{point}^{sc}$ (Equation \ref{eqn:point_triplet}), where:
        \begin{align}
            \mathcal{L}_{point}^{cc} =& \frac{1}{N_{pose}}\sum_{j=1}^{N_{pose}} (m + d(F(\mathbf{w}^{A1})_j, F(\mathbf{x}^{A2})_j, F(\mathbf{x}^{A2})_{k}) )^{+}, \\
            \begin{split}
                \mathcal{L}_{point}^{sc} =& \frac{1}{N_{id}}\sum_{j=1}^{N_{id}} (m + d(F(\mathbf{w}^{B1})_j, F(\mathbf{x}^{B3})_j, F(\mathbf{x}^{B3})_{k}) )^{+} \\
                & + \frac{1}{N_{pose}}\sum_{j=1}^{N_{pose}} (m + d(F(\tilde{\mathbf{w}}^{A1})_j, F(\mathbf{x}^{A2})_j, F(\mathbf{x}^{A2})_{k}) )^{+}.
            \end{split}
        \end{align}
	    \STATE \label{alg:line:unsup_end} Compute gradients and update:
            \begin{equation}
                \theta_G \leftarrow \theta_G - \eta_t \frac{\partial \mathcal{L}_{U}}{\partial \theta_G},
            \end{equation}
        where
        \begin{equation}
            \mathcal{L}_{U} = \mathcal{L}_{cc} + \mathcal{L}_{sc} + \mathcal{L}_{mesh}^{cc} + \mathcal{L}_{mesh}^{ss} + \mathcal{L}_{point}.
        \end{equation}
        \STATE $t\leftarrow t+1$.
	    \UNTIL{$t=t_{max}$.}
	\end{algorithmic}
\end{algorithm*}

\begin{algorithm}
\caption{\emph{MAPConNet in Semi-supervised Learning.}}
\label{alg:semi}
	\begin{algorithmic}[1]
	    \REQUIRE Labelled (pose aligned) dataset
	    $$\mathcal{X}^L = \{\mathbf{x}^{A,1}, \ldots, \mathbf{x}^{A,n}, \mathbf{x}^{B,1}, \ldots, \mathbf{x}^{B,n}, \ldots\},$$
	    and unlabelled (pose unaligned) dataset
	    $$\mathcal{X}^U = \{\mathbf{x}^{\alpha, \alpha_1}, \ldots, \mathbf{x}^{\alpha, \alpha_{n_{\alpha}}}, \mathbf{x}^{\beta, \beta_1}, \ldots, \mathbf{x}^{\beta, \beta_{n_{\beta}}},\ldots\},$$
	    where $\alpha,\beta,\ldots$ are used to distinguish identities in $\mathcal{X}^U$ from those in $\mathcal{X}^L$, and $\alpha_1,\beta_1,\ldots$ are used to emphasize that poses are unaligned across identities.
	    \STATE \textbf{Initialise:} Generator $G$ with trainable weights $\theta_G$, feature extractor $F$ (which is a part of $G$), time step $t=0$, initial learning rate $\eta_0$, maximum time step $t_{max}$, and current number of unlabelled iterations done $n_U=0$.
	    \REPEAT
	    \IF{iteration $t$ is labelled}
	        \STATE Sample a triplet from $\mathcal{X}^L$, e.g. $(\mathbf{x}^{A1}, \mathbf{x}^{B2}, \mathbf{x}^{B1})$, in the same format as line \ref{alg:line:sup_data} of Algorithm \ref{alg:sup}
	        \STATE Execute lines \ref{alg:line:sup_start}--\ref{alg:line:sup_end} of Algorithm \ref{alg:sup}.
	    \ELSE
	        \IF{$n_U\equiv 0 \pmod{3}$}
	            \STATE Sample a triplet from both $\mathcal{X}^L$ and $\mathcal{X}^U$, e.g. $\mathbf{x}^{\alpha,\alpha_1},\mathbf{x}^{\alpha,\alpha_2}\in\mathcal{X}^U$ and $\mathbf{x}^{C,3}\in\mathcal{X}^L$.
	        \ELSIF{$n_U\equiv 1 \pmod{3}$}
	            \STATE Sample a triplet from both $\mathcal{X}^L$ and $\mathcal{X}^U$, e.g. $\mathbf{x}^{D,4},\mathbf{x}^{D,5}\in\mathcal{X}^L$ and $\mathbf{x}^{\beta,\beta_1}\in\mathcal{X}^U$.
	        \ELSE
	            \STATE Sample a triplet from $\mathcal{X}^U$ only, e.g. $\mathbf{x}^{\gamma,\gamma_1},\mathbf{x}^{\gamma,\gamma_2},\mathbf{x}^{\delta,\delta_1}\in\mathcal{X}^U$.
	        \ENDIF
	        \STATE With a triplet in the same format as line \ref{alg:line:unsup_data} of Algorithm \ref{alg:unsup}, execute lines \ref{alg:line:unsup_start}--\ref{alg:line:unsup_end} of Algorithm \ref{alg:unsup}.
            \STATE $n_U\leftarrow n_U+1$.
	    \ENDIF
	    \STATE $t\leftarrow t+1$.
	    \UNTIL{$t=t_{max}$.}
	\end{algorithmic}
\end{algorithm}

\section{Additional results}
\begin{figure*}
\begin{center}
\includegraphics[width=0.99\linewidth]{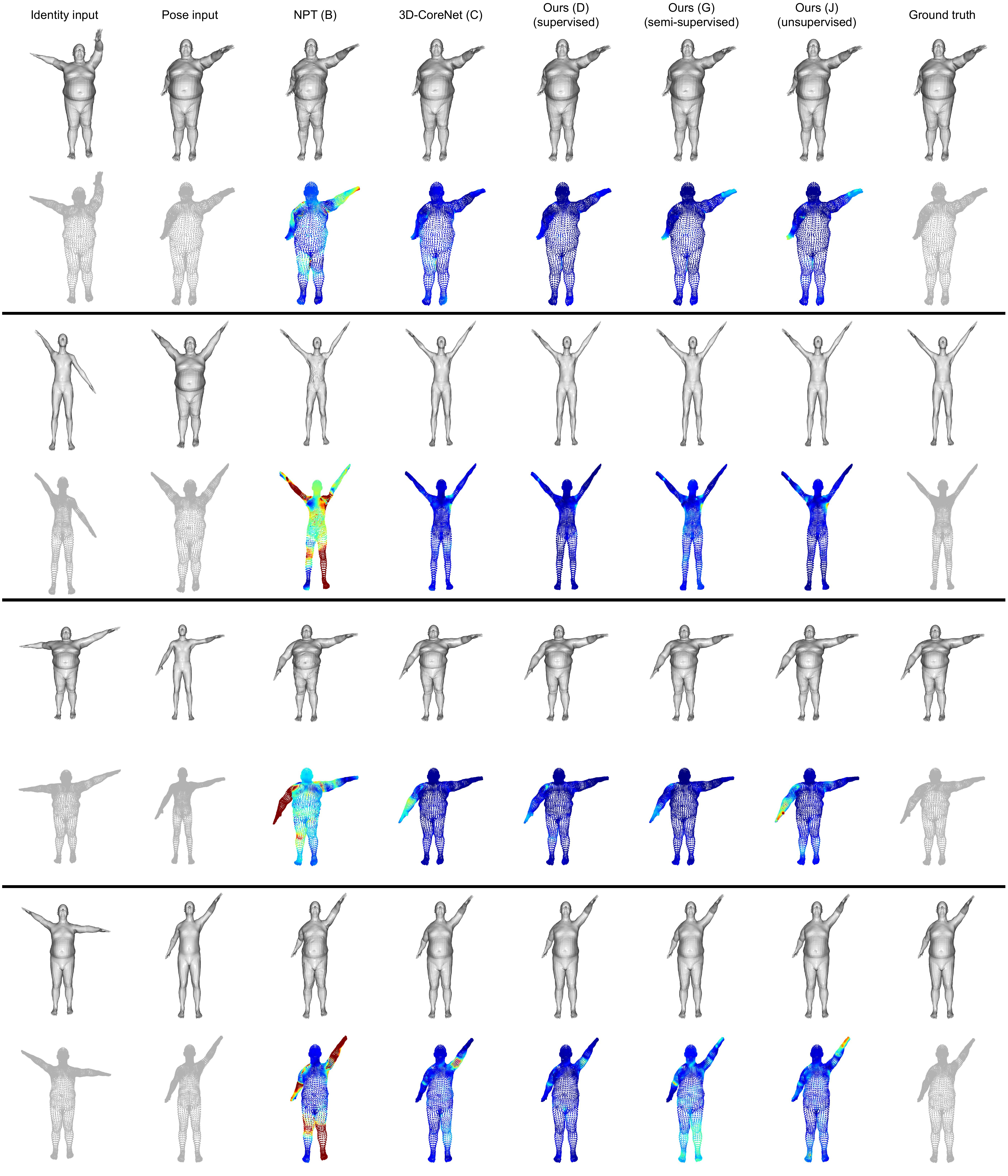}
\end{center}
\caption{\textbf{Additional qualitative comparison on unseen SMPL inputs.} These are additional examples of pose transfer performed using various methods (labels defined in Table \ref{tab:quantitative_3d}). Similar to Figure \ref{fig:qualitative_3d}, the first rows show the rendered surfaces and the second rows show the corresponding point clouds with PMD heatmaps (dark red: high error; dark blue: low error).}
\label{fig:qualitative_additional_smpl}
\end{figure*}

\begin{figure*}
\begin{center}
\includegraphics[width=0.99\linewidth]{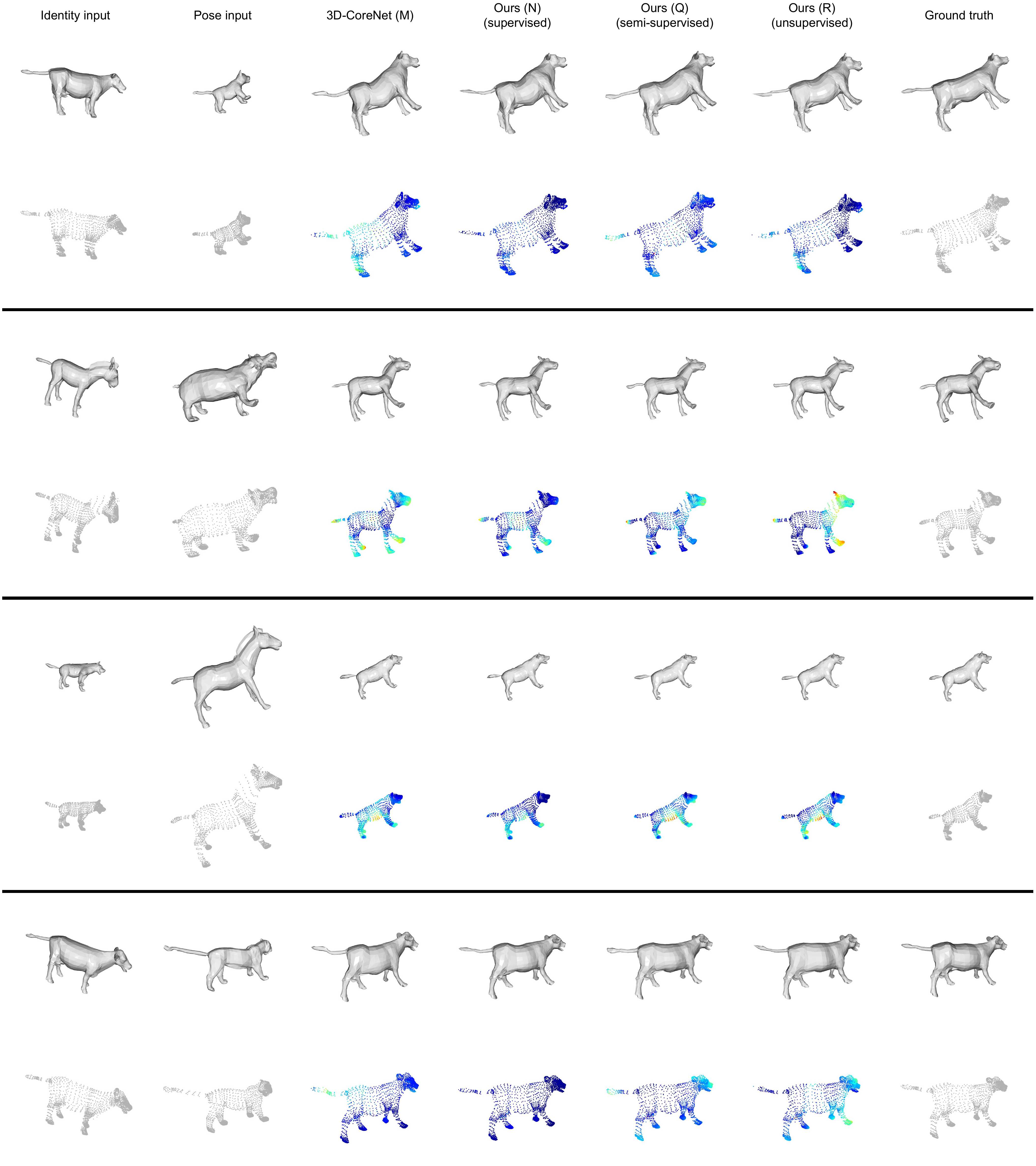}
\end{center}
\caption{\textbf{Additional qualitative comparison on unseen SMAL inputs.} These are additional examples of pose transfer performed using various methods (labels defined in Table \ref{tab:quantitative_3d}). Similar to Figure \ref{fig:qualitative_animal}, the first rows show the rendered surfaces and the second rows show the corresponding point clouds with PMD heatmaps (dark red: high error; dark blue: low error).}
\label{fig:qualitative_additional_smal}
\end{figure*}

\begin{figure*}
\begin{center}
\includegraphics[width=0.99\linewidth]{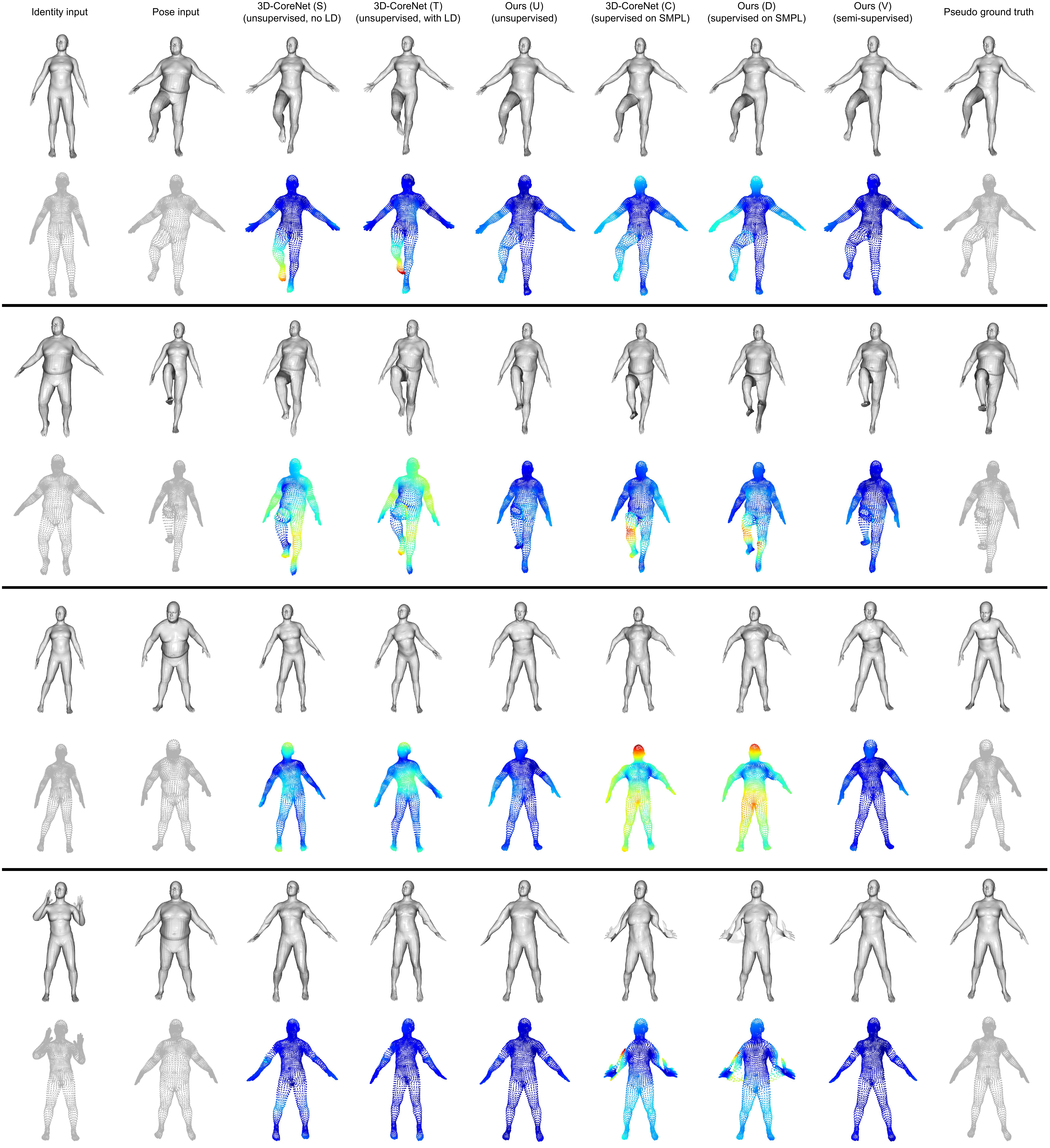}
\end{center}
\caption{\textbf{Additional qualitative comparison on unseen DFAUST inputs.} These are additional examples of pose transfer performed using various methods (labels defined in Table \ref{tab:quantitative_3d}). The first rows show the rendered surfaces and the second rows show the corresponding point clouds with PMD heatmaps (dark red: high error; dark blue: low error). The pseudo ground truths are generated using SMPL+H.}
\label{fig:qualitative_additional_dfaust}
\end{figure*}

\begin{figure*}
\begin{center}
\includegraphics[width=0.99\linewidth]{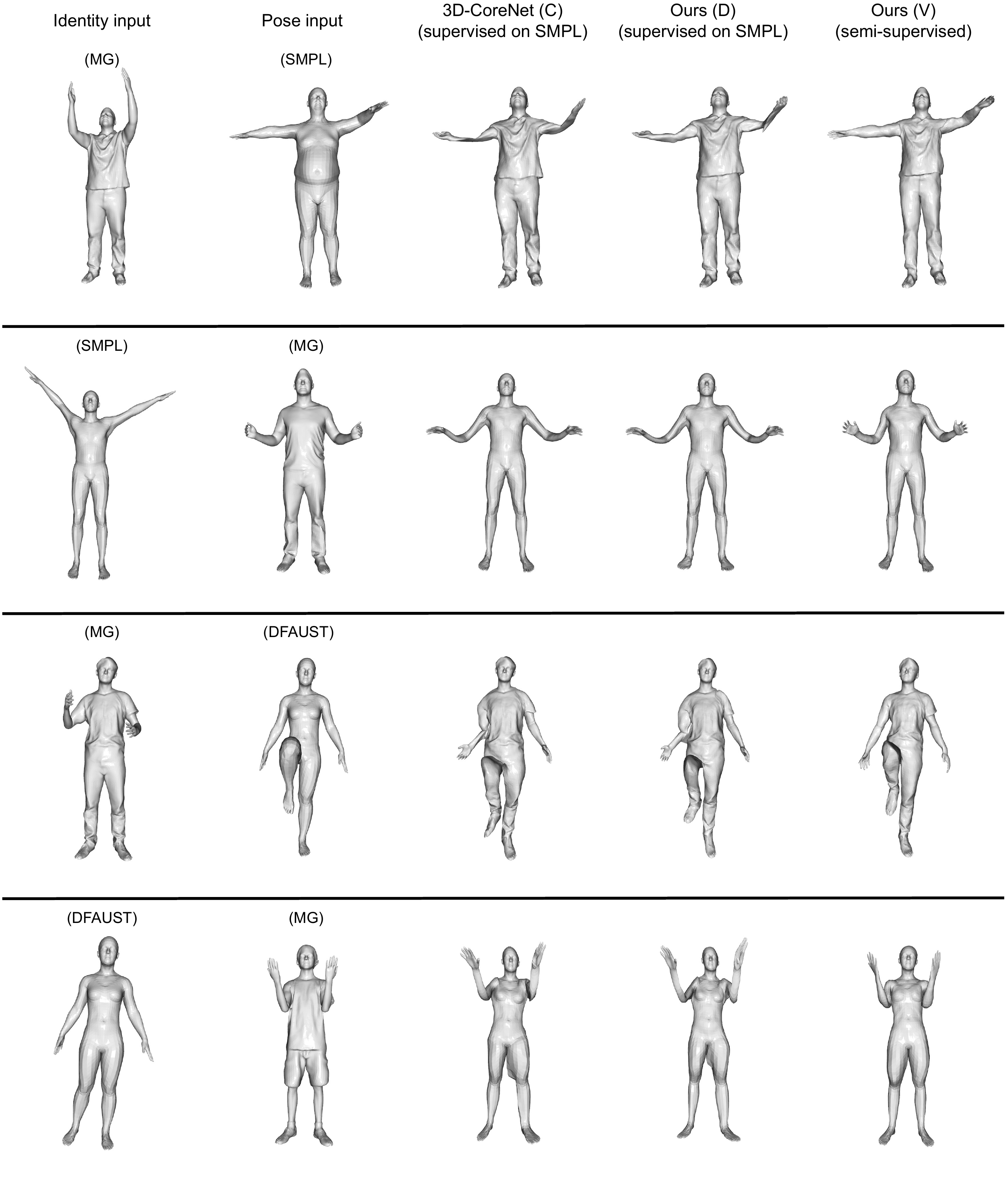}
\end{center}
\caption{\textbf{Additional qualitative comparison on unseen MG, SMPL, and DFAUST inputs.} These are additional examples of pose transfer performed using various methods (labels defined in Table \ref{tab:quantitative_3d}) with inputs from a mixture of different datasets (source dataset labelled in parentheses).}
\label{fig:qualitative_additional_mg}
\end{figure*}

We include additional qualitative comparisons on unseen input meshes from SMPL (see Figure \ref{fig:qualitative_additional_smpl}), SMAL (see Figure \ref{fig:qualitative_additional_smal}), DFAUST (see Figure \ref{fig:qualitative_additional_dfaust}), and MG (see Figure \ref{fig:qualitative_additional_mg}). Similar to observations we obtained from Figures \ref{fig:qualitative_3d} and \ref{fig:qualitative_animal}, our {\modelname} while under a fully supervised setting can generate more accurate outputs compared to 3D-CoreNet, particularly in the limbs, as shown by the PMD heatmaps. Our unsupervised and semi-supervised results are also comparable to the supervised ones.

In Figure \ref{fig:qualitative_additional_dfaust}, one can clearly observe that our unsupervised and semi-supervised models (\emph{U}) and (\emph{V}) produce more accurate transfer results on DFAUST inputs than supervised models (\emph{C}) and (\emph{D}) that can only be trained using labelled data SMPL and partial models (\emph{S}) and (\emph{T}). In Figure \ref{fig:qualitative_additional_mg}, the identity and pose inputs are chosen from different datasets but one of them is from MG. In this scenario, the model has to not only perform pose transfer on unseen meshes across different domains, but also handle the discrepancy in the numbers of vertices of both inputs. Our semi-supervised model (\emph{V}) again produces more realistic outputs than the supervised models, demonstrating its generalisability to complex unseen topologies.

\begin{figure*}
\begin{center}
\includegraphics[width=0.99\linewidth]{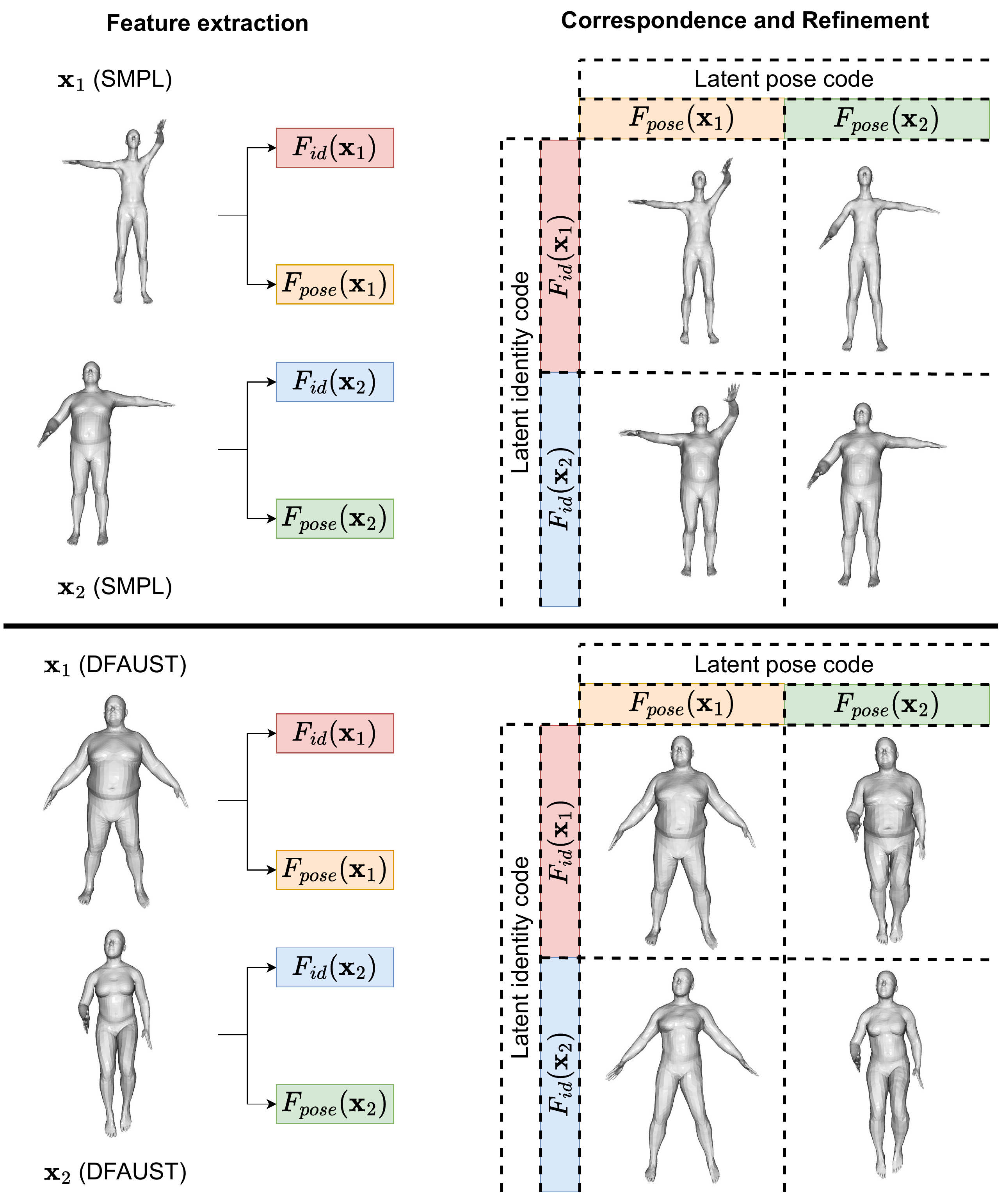}
\end{center}
\caption{\textbf{Outputs generated from disentangled latent codes.} These results are generated by method (\emph{V}) defined in Table \ref{tab:quantitative_3d} using various combinations of disentangled latent representations as described in Section \ref{ssec:lat_dis}. The vertex orders of $\mathbf{x}_1$ and $\mathbf{x}_2$ are aligned in order for their latent codes to be combined and processed by the model correctly.}
\label{fig:latent}
\end{figure*}

In addition, Figure \ref{fig:latent} shows outputs generated using various combinations of our disentangled latent representations. As one can observe, our model can indeed produce latent representations that are disentangled into identity and pose. This disentanglement is achieved in both synthetic and realistic datasets.

\begin{figure*}
\begin{center}
\includegraphics[width=0.99\linewidth]{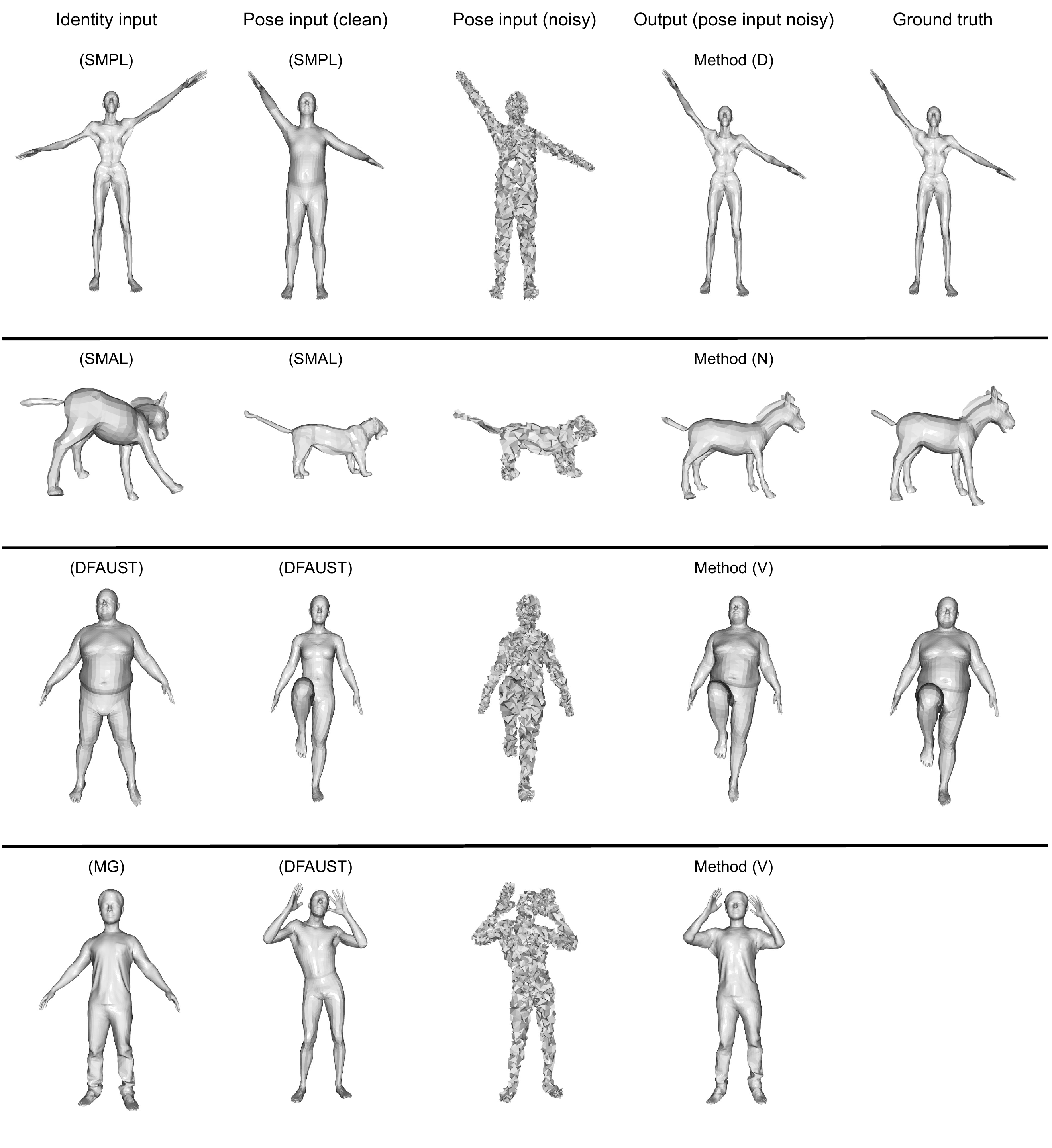}
\end{center}
\caption{\textbf{Outputs generated from noisy inputs.} These results are generated from a noisy pose input using various methods (labels defined in Table \ref{tab:quantitative_3d}). The vertices of the pose input are perturbed using random uniform noises. The clean pose inputs are shown here for reference but are not used as inputs for the model.}
\label{fig:qualitative_noise}
\end{figure*}

Finally, Figure \ref{fig:qualitative_noise} shows outputs generated from noisy inputs. A small but significant random uniform noise is added to each vertex of the pose input. As shown by the figure, out models are still able to generate outputs with the accurate pose and identity across multiple datasets.

\section{Limitations and future work}
Whilst our method achieves state-of-the-art results in our experiments, there are still limitations to be addressed in future work. For instance, we require both pose and identity inputs in CC during unsupervised learning to have the same vertex order, as the pose input is used as ground truth to supervise the model output which also needs to have the same vertex order as the identity input. There are potential solutions, such as modifying the behaviour of the correspondence module, or designing order-invariant loss functions (alternative to $\mathcal{L}_{rec}$) without impacting correspondence learning. However, this is not a serious issue since it is not impractical to maintain the same ordering within each identity during real world motion capture.

In semi-supervised learning, as our model can utilise unlabelled samples, it achieves superior performance compared to prior supervised methods which only have access to a limited number of labelled samples. Our model also tolerates a certain degree of domain gap between the labelled and unlabelled datasets as our model (\emph{V}) can handle unseen inputs from both SMPL and DFAUST. However, this might not necessarily hold if the labelled and unlabelled sets are from \emph{drastically} different domains, for instance, humans and animals. This might require fundamentally redesigning the correspondence module for cross-domain pose transfer. This is also a challenging problem to tackle due to the lack of cross-domain ground truths such as output, correspondence, or template pose.

\end{document}